\documentclass[10pt]{wlscirep}
\usepackage[utf8]{inputenc}
\usepackage[T1]{fontenc}
\usepackage{bm}
\usepackage{siunitx}
\usepackage{algorithm}
\usepackage{algpseudocode}
\usepackage{threeparttable}
\usepackage{adjustbox}

\usepackage{float}

\definecolor{algblue}{HTML}{2F6690}
\definecolor{algblueback}{HTML}{EEF5FA}
\definecolor{algteal}{HTML}{3A7D78}
\definecolor{algtealback}{HTML}{EDF7F5}
\definecolor{alggold}{HTML}{A66A20}
\definecolor{alggoldback}{HTML}{FFF7E8}
\definecolor{algslate}{HTML}{596775}
\definecolor{algslateback}{HTML}{F2F4F6}
\newcommand{\AlgPhase}[3]{%
    \Statex \vspace{2pt}%
    \begingroup
    \setlength{\fboxsep}{3pt}%
    \setlength{\fboxrule}{0.45pt}%
    \fcolorbox{#1}{#2}{%
        \parbox{\dimexpr\linewidth-2\fboxsep-2\fboxrule\relax}{%
            \color{#1}\sffamily\bfseries\small #3%
        }%
    }%
    \endgroup
    \vspace{1pt}%
}
\algrenewcommand\algorithmicindent{1.25em}
\algrenewcommand\algorithmicrequire{\textbf{Input:}}
\algrenewcommand\algorithmicensure{\textbf{Output:}}

\floatstyle{plain}
\newfloat{extfigure}{tbp}{lofex}
\floatname{extfigure}{Extended Data Figure}
\title{Minute-Scale Training for Microrobot Navigation}

\author[1]{Yinghan~Sun}
\author[1]{Aoji~Zhu}
\author[2]{Xiang~Ji}
\author[1]{Yamei~Li}
\author[1]{Jiachi~Zhao}
\author[1]{Yun~Wang}
\author[2,3,*]{Li~Zhang}
\author[4,*]{Huijun~Gao}
\author[1,5,6,*]{Lidong~Yang}

\affil[1]{State Key Laboratory of Ultra-precision Machining Technology, Department of Industrial and Systems Engineering, The Hong Kong Polytechnic University, Hong Kong, China}
\affil[2]{Department of Mechanical and Automation Engineering, The Chinese University of Hong Kong, Hong Kong, China}
\affil[3]{Shenzhen Loop Area Institute, Shenzhen, China}
\affil[4]{Research Institute of Intelligent Control and Systems, Harbin Institute of Technology, Harbin, China}
\affil[5]{Research Institute for Advanced Manufacturing, The Hong Kong Polytechnic University, Hong Kong, China}
\affil[6]{PolyU-PUTH Medicine-Engineering Collaborative Innovation Research Laboratory, The Hong Kong Polytechnic University, Hong Kong, China}
\affil[*]{Corresponding authors. Email: lizhang@cuhk.edu.hk; hjgao@hit.edu.cn; lidong.yang@polyu.edu.hk}

\begin{abstract} \bfseries \boldmath
Microrobots hold significant potential for various applications, where targeted navigation is a basic requirement. Deep reinforcement learning (DRL) has recently emerged as a powerful paradigm for fully autonomous microrobot navigation. Yet, current DRL-based approaches pay limited attention to learning efficiency and effectiveness, requiring hours to days for model training. Consequently, this impedes both rapid practical deployment and parameter optimization. To address these challenges, we present a learning framework that enables effective microrobot navigation policies to be trained within minutes. In the proposed framework, we develop a fully vectorized simulator with more than 10,000 artificial vascular environments, parallelizing dynamics, LiDAR-inspired perception, and feasibility checks across thousands of environments to achieve roughly 190,000 transitions per second. To achieve effectiveness in fast training, we propose a task-shaping-regularization (TSR) reward framework. The TSR framework accelerates convergence, improves final performance, reduces action variation by at least 33.7\%, and increases obstacle clearance by at least 2.1\% across all evaluated scenarios. Results show that the proposed learning framework reduces training time to under 10 minutes, while supporting zero-shot deployment across distinct microrobot types and navigation scenarios. Collectively, this framework can substantially shorten the design loop and accelerate the deployment of autonomous microrobots.
\end{abstract}

\begin{document}

\flushbottom
\maketitle

\thispagestyle{empty}

\section{Introduction}

Microrobots represent a transformative frontier in robotics \cite{app_drug_delivery_2023science, sitti2015biomedical, iacovacci2024medical, zhang2019robotic}, providing wireless and non-invasive access to confined regions for precise operations beyond the capabilities of conventional robots. 
Among various actuation methods, magnetic actuation has become a widely adopted approach for microrobot control~\cite{xu2015magnetic, yang2021motion}, favored for its strong penetration capability, high biocompatibility, and real-time controllability. 
Researchers have demonstrated the considerable potential of microrobotic systems across a wide range of biomedical applications, including targeted drug delivery~\cite{app_drug_delivery_2022acsnano, app_targeted_drug_delivery_2025science}, precision localized therapy~\cite{app_precision_local_therapy_2022sa, app_plt_dt_2022sa, app_selective_embolization_2022sa}, micromanipulation~\cite{xu2024survey, jiang2024automated}, and other biomedical tasks~\cite{li2017micro}. 
Among them, targeted delivery and therapy in vascular systems are of particular interest~\cite{del2023ultrasound, wang2022adaptive, go2023soft}, for which microrobots should be capable of navigating accurately through confined and branched vascular environments with tortuous boundaries. 
Accordingly, substantial research efforts have been devoted to achieving autonomous navigation in such complex and unstructured settings~\cite{huan2022path, yang2022hierarchical}.

Autonomous control and navigation of microrobots have been studied extensively. 
Many existing microrobot navigation frameworks rely on classical planning and optimization methods, including artificial potential fields~\cite{fan2022obstacle}, search-based planners such as A*~\cite{astar}, sampling-based planners such as RRT~\cite{rrt}, and optimization-based methods. 
Classical planners, including A*~\cite{astar_2021tmech}, RRT~\cite{rrt_2019tii, rrt_2021tro, rrt_2022tase}, and their variants, can generate feasible trajectories from the current position to a target. 
However, their performance typically depends on global map priors, which limits applicability and generalization in unstructured and complex environments~\cite{astar_2021tmech}. 
In addition, the latency introduced by repeated replanning can preclude real-time operation~\cite{dqn_2d_time_penalty_2024engineering}. 
Optimization-based methods offer an alternative by casting navigation as a constrained optimization problem that encodes task objectives and system constraints~\cite{oc_ldyang_2020tase, oc_2023tro}. 
Yet such methods require tractable mathematical descriptions of the environment, which are difficult to obtain in unstructured settings. As a result, they are usually restricted to structured scenarios without strongly dynamic features.

\begin{figure}[htbp]
    \centering
    \includegraphics[width=\textwidth, trim = 0 0 0 0, clip]{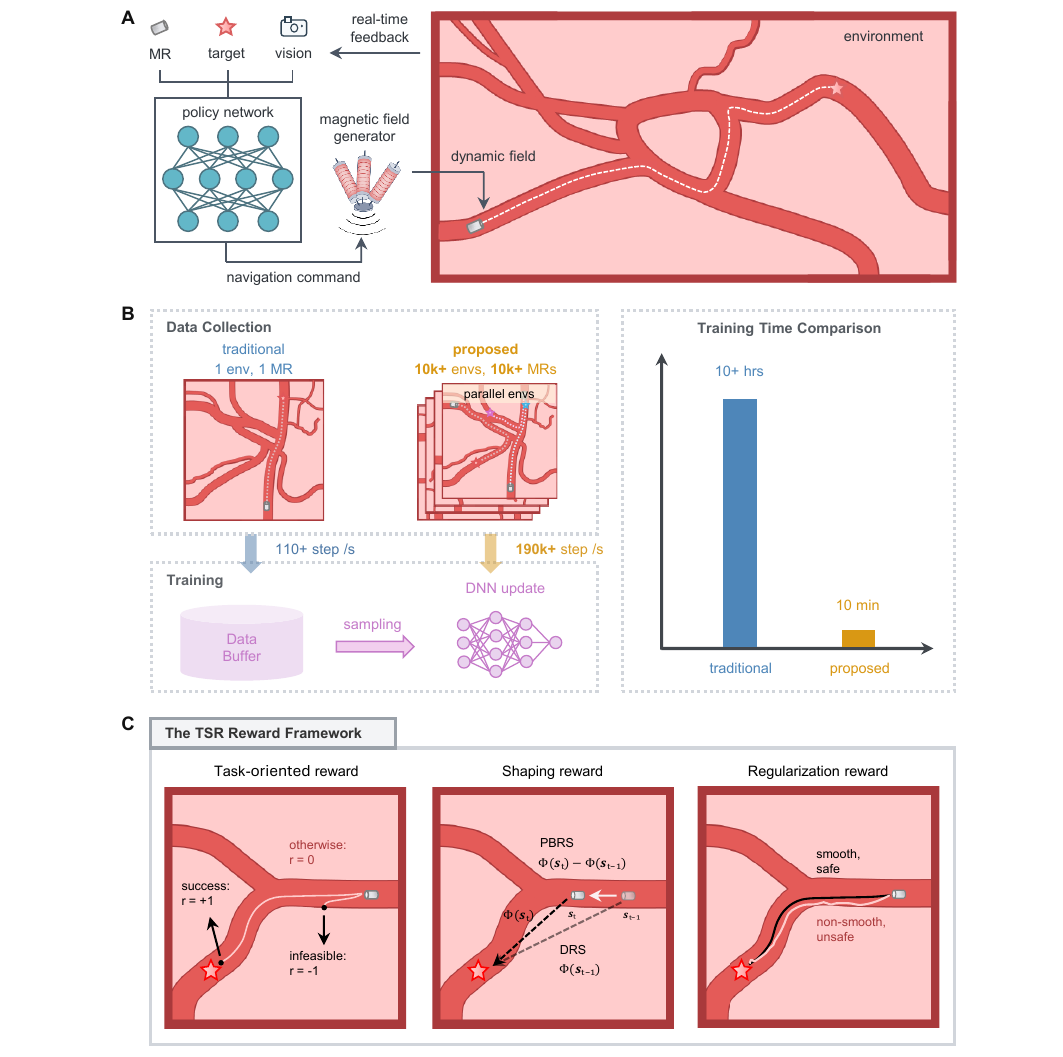}
    \caption{\textbf{Learning autonomous navigation skills for microrobots in minutes.} (A) Schematic of learning-based autonomous microrobot (MR) navigation. 
    (B) Built on a large-scale vascular dataset, our simulator achieves substantially higher throughput, enabling minute-scale training for microrobot navigation.
    (C) Schematic of the task-shaping-regularization (TSR) reward framework. It promotes fast convergence and high-quality trajectories: task-oriented reward encodes sparse success and failure signals; shaping reward densifies feedback through stepwise progress toward the target, implemented as a unit-discount potential-based shaping term ($\gamma_s = 1$); and regularization reward encourages smooth and collision-free motion.
    }
    \label{fig:teaser}
\end{figure}

In recent years, to address these limitations, Deep Reinforcement Learning (DRL)~\cite{dqn_2015nature} has emerged as a promising data-driven paradigm without reliance on precise modeling~\cite{ldyang_ml_microrobot_review_2024nmi}.
By interacting with the environment, microrobots can learn adaptive policies for navigating complex geometric structures~\cite{mbrl_2025nmi} and dynamic physical conditions~\cite{rl_soft_flow_2023tcyber, wang2024deep}. 
Researchers have applied DRL to diverse tasks, including target reaching~\cite{rl_real_target_reaching_2023, 3d_4stage_2024nmi}, channel traversal~\cite{mbrl_2025nmi, dqn_2d_time_penalty_2023icra, dqn_2d_time_penalty_2024engineering}, obstacle avoidance~\cite{dqn_2d_time_penalty_2023icra, dqn_2d_time_penalty_2024engineering}, and collaborative task planning~\cite{rl_collaborative_task_planning_2025tmech}. 
However, these advances often come with substantial training costs. 
For example, Salehi et al.~\cite{real_2024micromachines} required 45 hours of on-hardware training to learn a target-reaching policy. 
Abbasi et al.~\cite{3d_4stage_2024nmi} proposed a four-stage sim-to-real pipeline that still required 10 hours of simulation and 4 hours of hardware training, even for 2D tasks. 
Medany et al.~\cite{mbrl_2025nmi} used model-based reinforcement learning (RL), with model convergence taking up to 10 days. 
Such training times hinder algorithmic exploration, systematic hyperparameter tuning, and rapid iteration towards practical deployment.

These high training costs expose two critical, largely unaddressed challenges in the field: learning efficiency and learning effectiveness. By efficiency, we refer to obtaining high‑performing policies with minimal wall‑clock training time and high data throughput.
To date, most existing methods simulate only a single environment instance and generate at most one transition per timestep~\cite{dqn_2d_time_penalty_2023icra, dqn_2d_time_penalty_2024engineering, 3d_4stage_2024nmi}, resulting in low throughput and extended training. 
By effectiveness, we mean rapid convergence and high‑quality navigation behaviour within a fixed computational budget. 
This property is strongly shaped by the problem formulation, in particular, the reward design, which specifies the optimization objective and ultimately governs the learned policy~\cite{sutton_rl_book}. 
Yet the existing literature offers little systematic analysis of how different reward choices affect learning effectiveness.

In this work, we address both challenges by introducing a training framework that enables microrobot navigation policies to be trained within minutes in complex, unstructured environments (Fig.~\ref{fig:teaser}).
To achieve this, we systematically investigate the key factors underlying efficient and effective policy learning.

To enable efficient learning, we develop a fully vectorized simulator based on a large-scale vascular dataset comprising more than 10,000 artificial vascular channels. 
These channels are generated to capture key characteristics of real vascular environments, including variations in diameter, branched structures, and open-sided geometries with curved boundaries. 
Although vectorized simulation is well established in reinforcement learning, massively parallel simulation for microrobot navigation in complex vascular environments remains underexplored, and its feasibility and performance benefits have yet to be systematically demonstrated.
To address this gap and establish a methodological foundation for vectorized microrobot learning, we formulate the vectorized strategy for computing microrobot dynamics, together with a vectorized LiDAR-inspired visual feature extractor under irregular geometric constraints.
This design allows the simulation of thousands of environments in parallel and achieves a throughput of approximately 190,000 transitions per second (Fig.~\ref{fig:teaser}(B)). 
Consequently, we can train a navigation policy within 10 minutes and deploy it in a zero‑shot manner on different types of microrobots across diverse scenarios.

To achieve effective learning, reward design should be systematically investigated.
We propose a task–shaping–regularization (TSR) reward framework, as illustrated in Fig.~\ref{fig:teaser}(C).
To provide a reward design methodology for microrobot learning, our study reveals the distinct functional roles of the dense reward components: the shaping term mainly improves convergence speed, whereas the regularization term mainly improves the quality of the learned behaviour. 
Following this design framework, researchers can tailor alternative reward terms for different microrobots and application scenarios.

Leveraging these components, we show that the trained policy can be transferred to distinct microrobotic platforms in a zero-shot manner. 
These policies can navigate different microrobots (e.g., a pollen‑based microparticle and a helical microrobot) through vascular channels and dense fields of dynamic obstacles that are unseen during training, without any retraining or on‑hardware adaptation. 
We further embed the learned planar policy in hybrid controllers for navigation in an open 3D workspace and a confined vascular phantom. 
Overall, by enabling minute‑scale policy training, systematic reward analysis, and zero‑shot deployment on real microrobots, our framework can shorten the design loop for autonomous microrobotic navigation and accelerate deployment across diverse applications.
These advances represent a step towards intelligent microrobotic systems and extend machine intelligence to small-scale robotic platforms.

\begin{figure}[htbp]
    \centering
    \includegraphics[width=\textwidth, trim = 0 0 0 0, clip]{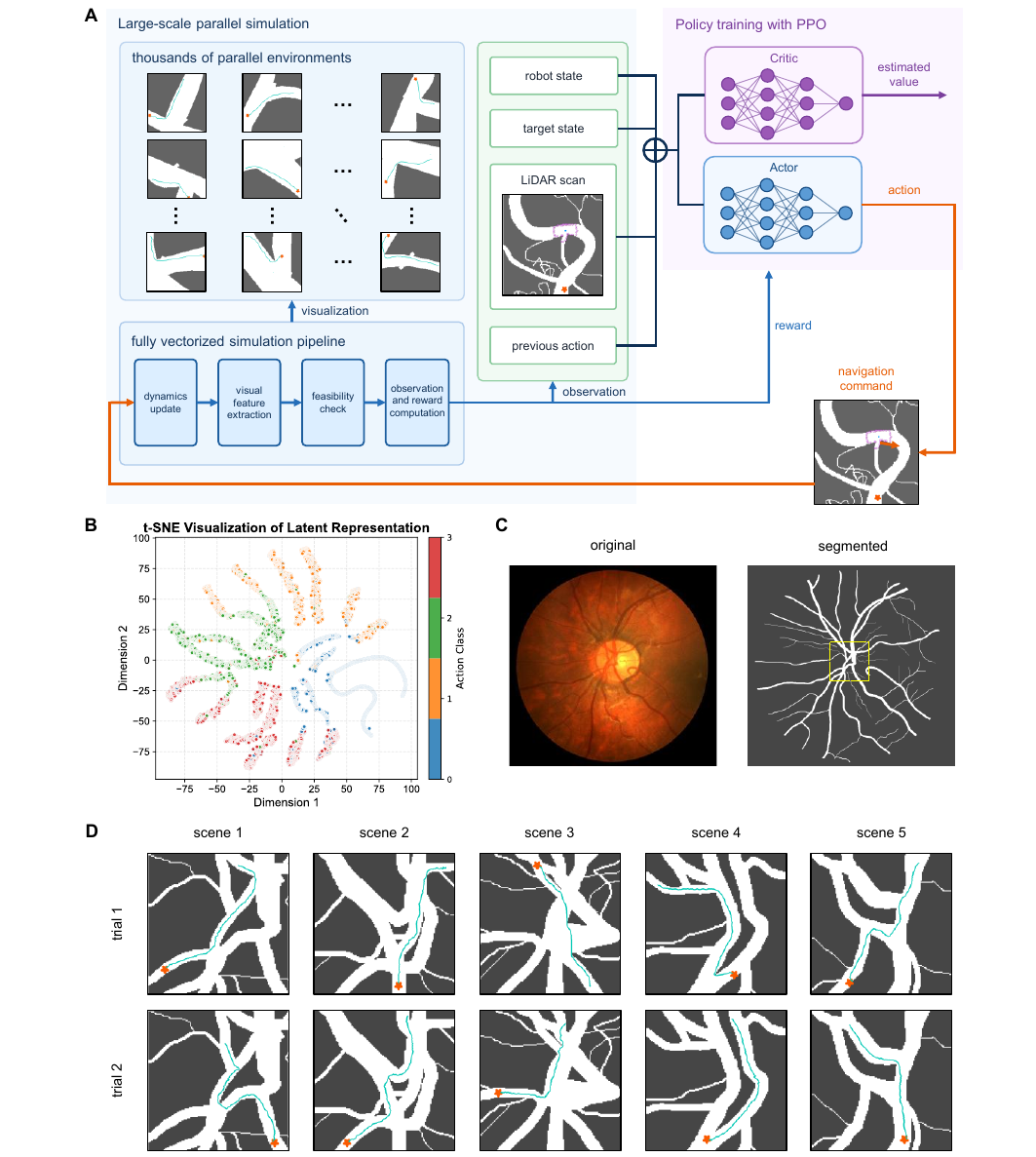}
    \caption{\textbf{Detailed illustration of the proposed framework.} (A) Illustration of the large-scale training. (B) t-SNE visualization of the latent representations of observations and their corresponding discretized actions collected from multiple navigation trajectories, where each point denotes a state representation. (C) Zero-shot transfer of the trained policy to unseen scenarios. As a representative example, we deploy the policy in vascular environments extracted from a publicly available retinal fundus image database. (D) Results for navigation in unseen environments.
    }
    \label{fig:method}
\end{figure}

\section{Results}

We consider an image-guided, closed-loop magnetic control system for microrobot navigation, in which external imaging systems provide real-time visual feedback (Fig.~\ref{fig:teaser}(A)). 
The policy network takes the microrobot state, target position, and visual observations as input, and outputs a navigation command, which is converted into the desired magnetic field parameters and executed by the magnetic actuation hardware in real time.
Our learning framework is shown in Fig.~\ref{fig:method}(A). 
The simulation environment is built on a large-scale artificial vascular dataset \cite{yang_swarm_2022nmi}. 
To improve computational efficiency, we vectorize the key components of the pipeline, including dynamics updates, visual feature extraction, and feasibility checking. 
In addition, we introduce a virtual LiDAR module to efficiently encode the local geometric features surrounding the microrobot.
The navigation policy is trained using proximal policy optimization (PPO) \cite{ppo_arxiv2017}, chosen for its empirical stability and ease of implementation in continuous-control tasks. 
Formulation and implementation details are provided in Methods.

\subsection{Large-scale Simulation-based Training}

\begin{figure}[htbp]
    \centering
    \includegraphics[width=\textwidth, trim = 0 0 0 0, clip]{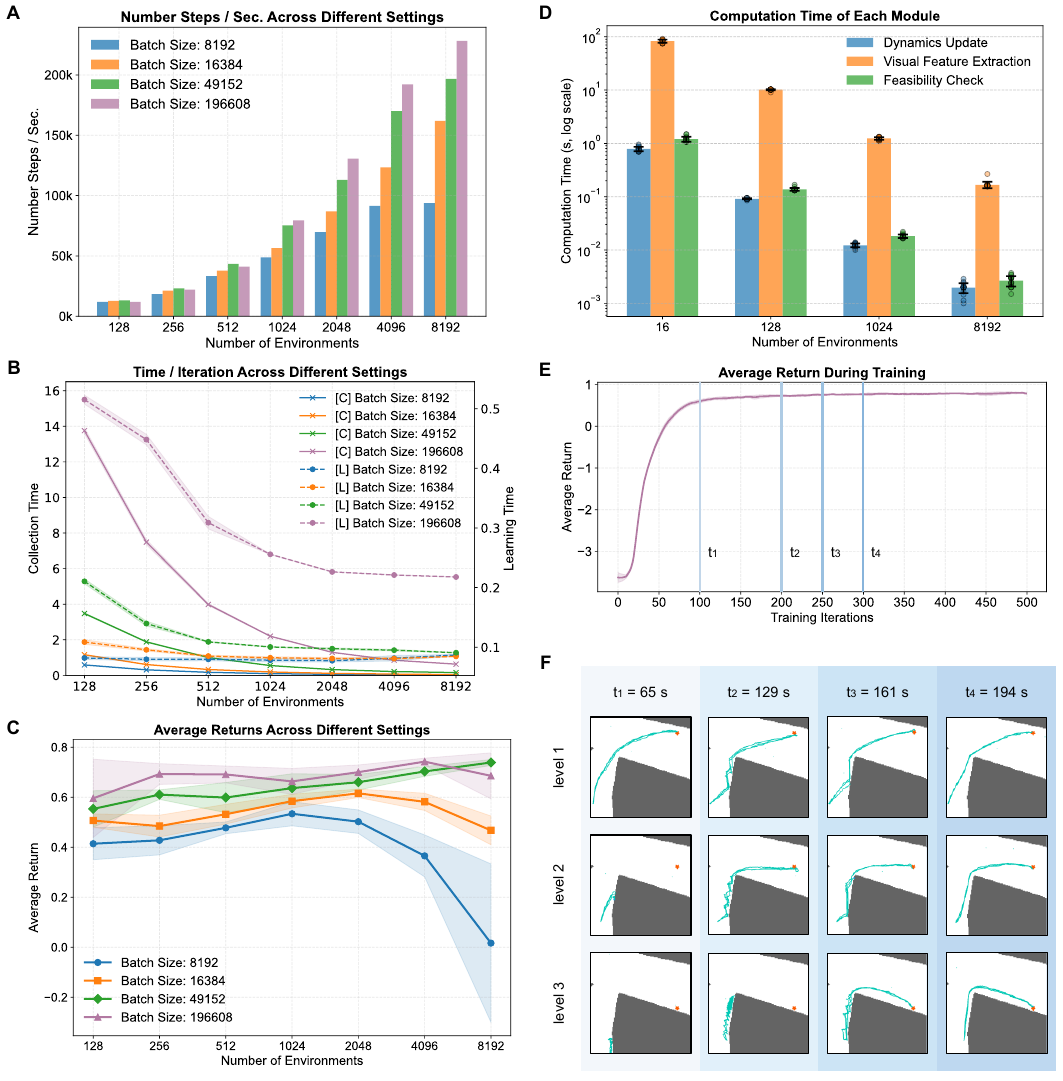}
    \caption{\textbf{Efficiency and effectiveness in the proposed large-scale training system.} (A) System throughput under different numbers of parallel environments and batch sizes, where each data point is averaged over ten independent runs. (B) Data-collection time and policy-learning time per training iteration across different hyperparameter settings. (C) Average return after 1,000 policy updates for different hyperparameter settings. (D) Computation time of each module at different parallelization levels, measured over 20 independent runs. Bars represent the arithmetic mean, and error bars indicate the standard deviation. (E) Training curve with 4,096 parallel environments; $t_1$ - $t_4$ mark four checkpoints used for visualization. 
    For (B)(C)(E), ten independent runs are conducted for each hyperparameter setting; solid lines indicate the mean performance and shaded regions denote the 95\% confidence interval across runs. We provide a more detailed discussion on the choice of statistical analysis in Supplementary Text 9. (F) Visualization of the training process using the policies at these four checkpoints. The policies are evaluated in a scenario with three difficulty levels, defined by the proximity of the initial and target positions to the vessel wall. The policy can solve the most challenging cases with excellent performance after only 194 seconds of training. Please see the Supplementary Movie S1.}
    \label{fig:results_large_scale_sim}
\end{figure}

Unlike conventional frameworks that typically simulate a single environment at a throughput of approximately 110 transitions per second \cite{dqn_2d_time_penalty_2023icra, 3d_4stage_2024nmi, mbrl_2025nmi}, our training pipeline leverages a large-scale vascular dataset containing 13,278 environments and concurrently simulates $N_\text{env}$ sampled environments during each training iteration.
The training is fully vectorized and executed as an iterative cycle with four core modules (detailed in Methods) at each time step, i.e., microrobot dynamics update, feasibility check, visual feature extraction, and observation and reward calculation.
As a result, the framework can process on the order of $10^5$ transitions per second, significantly reducing end-to-end training time from hours to minutes.

To understand how key hyperparameters shape large-scale training performance, we systematically quantify the training efficiency across two central hyperparameters: the batch size $B$ and the number of parallel environments $N_\text{env}$.
$B = N_{\text{env}} \times N_{\text{trans}}$, where $N_{\text{trans}}$ is the number of transitions collected per environment. 
Across a wide range of $(B, N_{\text{env}})$ combinations, we measure throughput in transitions per second (TPS), the time required for data collection and policy updates in each training iteration, and the average return after the same number of policy updates.

Fig.~\ref{fig:results_large_scale_sim}(A) shows that throughput generally increases with both $B$ and $N_\text{env}$. For example, 128 parallel environments with a batch size of 8,192 yield 11,815 TPS, whereas 8,192 parallel environments with a batch size of 196,608 reach 228,117 TPS, corresponding to a 19.3-fold speedup. However, excessively large batch sizes can reduce efficiency. When \(N_\text{env} \leq 512\), a batch size of 196{,}608 leads to lower throughput than smaller batches because each environment must generate long trajectories, which limits the benefit of parallel execution.

We also analyze the time costs for data collection and policy updates, as shown in Fig.~\ref{fig:results_large_scale_sim}(B). Increasing $N_\text{env}$ consistently reduces both components, especially at large batch sizes. 
For instance, with a batch size of 196{,}608, increasing \(N_\text{env}\) from 128 to 8{,}192 shortens the collection time from about \SI{16}{\second} to less than \SI{1}{\second}. The learning time exhibits a similar trend, indicating that large-scale parallelization not only accelerates rollout generation but also amortizes the cost of optimization over much larger batches.

To further assess the benefit of vectorization, we compare the computation time of each module under different levels of parallelism. As shown in Fig.~\ref{fig:results_large_scale_sim}(D), with a fixed batch size of 196,608, each eightfold increase in $N_\text{env}$ would reduce the computation time by approximately 7–10-fold at each tested increase in parallelism. 
Visual feature extraction is the dominant bottleneck at low parallelism, requiring nearly 100 s with 16 environments, but only about 0.2 s with 8,192 environments. 
Dynamics updates and feasibility checks are less costly overall, but they also benefit substantially from increased parallelism.

We finally examine whether the hyperparameter settings that maximize throughput also support effective learning. Fig.~\ref{fig:results_large_scale_sim}(C) shows that larger batch sizes generally yield higher returns. 
This can be explained by the PPO objective, which is estimated from sampled on-policy trajectories: small batches produce noisier estimates, increasing gradient variance and reducing optimization stability, whereas larger batches provide more reliable updates.
However, at a fixed batch size, excessive parallelization can degrade performance when too few transitions are collected from each environment. 
These results motivate using the largest batch size permitted by hardware constraints to fully exploit large-scale training; meanwhile, effective training requires balancing high parallelism with sufficient $N_\text{trans}$. 

Based on previous analysis and the available hardware, we adopt 4,096 parallel environments and a batch size of 196,608 in all subsequent experiments.
Under this setting, the policy converges rapidly. 
As shown in Fig.~\ref{fig:results_large_scale_sim}(E), the average return rises steeply at the beginning of training and quickly plateaus, indicating fast convergence and stable performance. 
Fig.~\ref{fig:results_large_scale_sim}(F) visualizes a representative scenario involving navigation around a sharp vessel corner,
evaluated at three difficulty levels defined by the distance between the initial/target positions and the vessel wall.
The policy first solves the easiest case, then the intermediate one, and finally the most difficult one with progressively smoother trajectories. Notably, even the most challenging case is solved after only 194 s of training. 
The extensive results demonstrate that the proposed large-scale simulation-based training framework enables minute-scale policy training while maintaining high success rates and stable navigation performance under the evaluated conditions.

\subsection{The TSR Reward Framework}

\begin{figure}[htbp]
    \centering
    \includegraphics[width=\textwidth, trim = 0 0 0 0, clip]{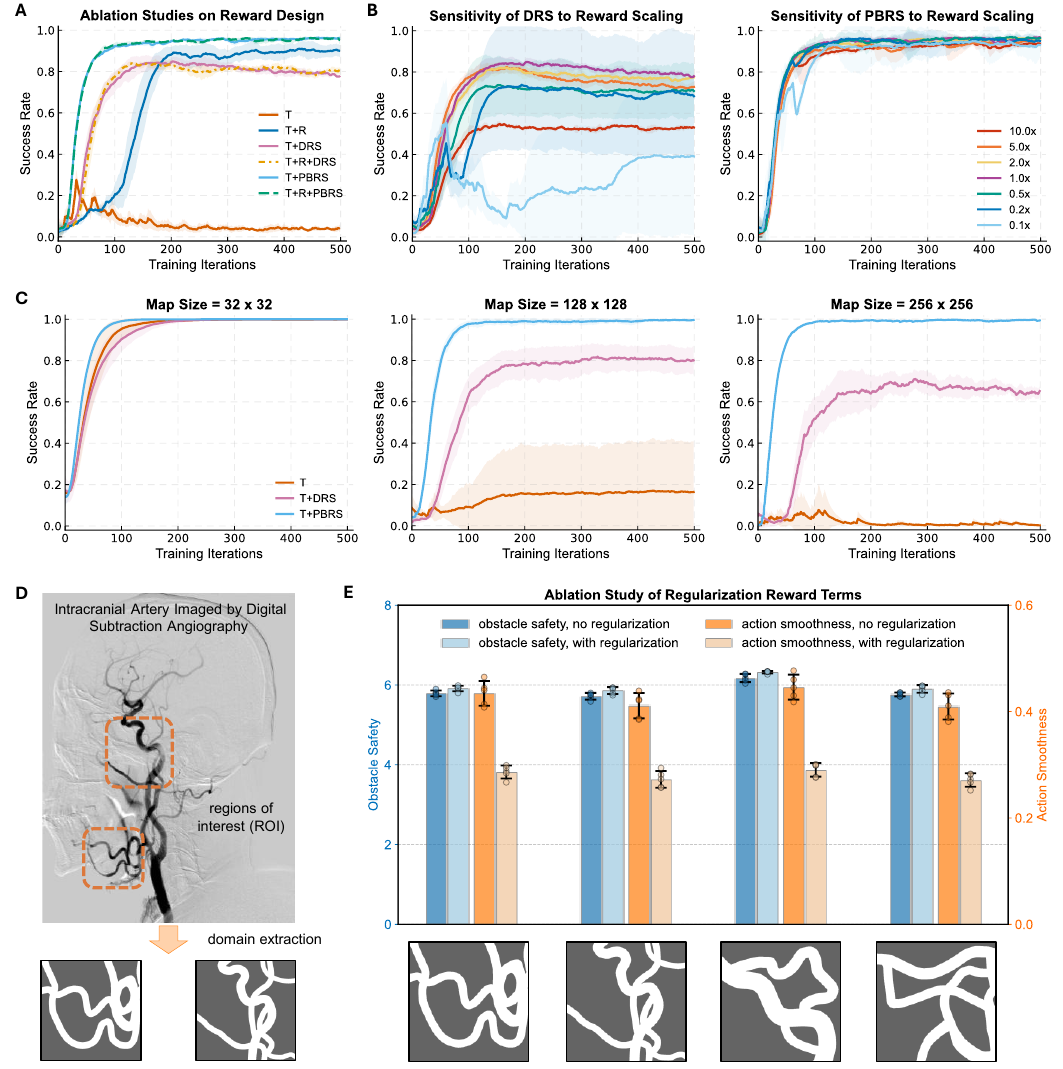}
    \caption{\textbf{Effectiveness of the components in the TSR reward framework.} For panels (A)-(C), ten independent runs are conducted for each experimental group; solid lines indicate the mean performance and shaded regions denote the 95\% confidence interval across runs. T denotes the task-oriented reward, R denotes the regularization reward, DRS denotes the direct reward shaping, and PBRS denotes the potential-based reward shaping. (A) Ablation study of the different components of the proposed TSR reward framework. (B) Sensitivity of the two shaping strategies to different reward scales; PBRS exhibits substantially more robust performance across reward scales. (C) Comparison of shaping strategies on free-space maps of varying sizes. PBRS remains robust across all map sizes, whereas DRS and the no-shaping baseline are sensitive to map size. (D) Illustration of the test scenarios, extracted from intracranial arteries imaged by digital subtraction angiography. (E) Ablation study of the regularization reward terms across four scenes. For each policy-scene combination, performance is averaged over 10,000 randomly initialized trajectories in each of five independent trials. Bars show the mean across trials, error bars indicate s.d., and points represent individual trial means. Higher obstacle-safety values and lower action-smoothness values indicate better performance.
    }
    \label{fig:results_reward_shaping}
\end{figure}

Reward design plays a pivotal role in RL, as it specifies the optimization objective that ultimately shapes both sample efficiency and the quality of the resulting behaviour.
Yet, for microrobot navigation, reward design has been driven largely by heuristic practice rather than systematic analysis. 
We therefore propose a task-shaping-regularization (TSR) reward framework (Fig.~\ref{fig:teaser}(C)) and clearly quantify the function of each component.
Detailed definitions and expressions of the three reward terms are included in Methods.

We first explore whether the task-oriented reward alone could support learning. As illustrated in Fig.~\ref{fig:teaser}(C), the task-oriented reward provides non-zero feedback only at episode termination, when the robot either reaches the goal or terminates because of a collision. 
Under this sparse supervision, training rarely converges reliably within 500 iterations across most settings (Fig.~\ref{fig:results_reward_shaping}(A)). 
The resulting policy is usable only on the smallest free-space map and deteriorates markedly as the free-space map size increases (Fig.~\ref{fig:results_reward_shaping}(C)). 
These results indicate that sparse task feedback alone is insufficient to guide efficient exploration in continuous microrobot navigation.

Dense intermediate rewards are often introduced through heuristic penalties, such as distance-to-goal terms~\cite{3d_4stage_2024nmi, mbrl_2025nmi} or time-cost penalties~\cite{rl_swarm_2021sr, dqn_2d_time_penalty_2023icra, dqn_2d_time_penalty_2024engineering}. However, the distinct roles of these terms have not been systematically characterized.
Here, we distinguish two classes of dense reward signals: shaping terms and regularization terms (Fig.~\ref{fig:teaser}(C)). 
Although both provide intermediate feedback, they serve different purposes. Shaping primarily improves credit assignment and accelerates convergence, whereas regularization primarily biases the learned policy towards desirable behaviours, such as smoother control and greater obstacle clearance, with a limited effect on convergence speed.

For reward shaping, we compare two representative strategies: direct reward shaping (DRS) and unit-discount potential-based reward shaping (PBRS). 
As shown in Fig.~\ref{fig:teaser}(C), DRS adds heuristic terms directly to the immediate reward, whereas PBRS derives the shaping signal from changes in a potential function between consecutive states. 
The distinction becomes clear when the negative Euclidean distance between the microrobot and the target is used as the potential. 
In DRS, using this distance directly as the immediate reward reflects only the current state and does not reveal whether the preceding action improves progress. For example, a reward of \(-10\) is identical whether the agent moves from a distance of 12 to 10 or from 8 to 10. 
By contrast, the proposed unit-discount PBRS evaluates the change in target-distance potential between consecutive states. It therefore rewards progress toward the target and penalizes movement away from it, thereby explicitly reflecting the consequence of the preceding action.
Consistent with this interpretation, PBRS provides a more informative learning signal and converges faster than DRS (Fig.~\ref{fig:results_reward_shaping}(A)).

We next evaluate robustness to the scaling of the shaping reward. 
The optimal weights for PBRS and DRS are first tuned by grid search, and the selected values are reported in Supplementary Fig.~S5. 
Using the optimal weights as references, we train policies with shaping weights ranging from \(0.1\times\) to \(10\times\), while keeping all other coefficients fixed. 
As shown in Fig.~\ref{fig:results_reward_shaping}(B), PBRS is markedly less sensitive to reward scaling: across all tested factors, it consistently achieves rapid convergence, strong final performance, and low inter-run variance. 
By contrast, DRS shows substantial variation in convergence speed, final performance, and variability across runs. 
This robustness makes PBRS easier to balance with other reward components.

We further test robustness across navigation tasks with different spatial scales. 
As shown in Fig.~\ref{fig:results_reward_shaping}(C), PBRS remains effective across all tested free-space map sizes, consistently achieving rapid convergence and high success rates with low variance. 
By contrast, both the task-only baseline and DRS become increasingly unstable as the map size increases. 
Notably, on the \(32\times 32\) free-space map, DRS converges more slowly than the task-only baseline despite reaching a similar final success rate, indicating that dense feedback is beneficial only when it encodes meaningful progress.

Regularization has a distinct effect. 
Across all four test scenarios in Fig.~\ref{fig:results_reward_shaping}(D), adding regularization consistently improves trajectory quality (Fig.~\ref{fig:results_reward_shaping}(E)), even when its influence on convergence speed is modest. 
Action smoothness increases by 34.0\%, 33.7\%, 34.9\%, and 33.7\% in the four test scenarios, respectively, and obstacle safety improves by 2.1\%, 2.5\%, 2.3\%, and 2.4\%, respectively. 
These gains indicate that regularization steers the policy towards dynamically smoother and safer behaviour that is not fully specified by the task objective alone.

Taken together, the ablation and robustness investigation reveals a clear division of labour within the TSR framework. 
The task-oriented reward specifies the goal of the behaviour, shaping rewards improve learnability by providing informative progress signals, and regularization rewards refine the final solution by favouring smoother and safer trajectories. 
This decomposition clarifies why dense rewards can have qualitatively different effects on policy learning, and it provides a principled methodology for reward design for microrobot navigation.

\subsection{Effective Visual Feature Encoding}
Given an observation comprising the microrobot and goal positions, LiDAR scan points, and the previous action, we investigate how the policy network encodes the information in its latent representation
and how this relates to the resulting action output.
Specifically, we extract the features from the final hidden layer of the policy network and visualize them using t-distributed stochastic neighbour embedding (t-SNE)~\cite{tsne_2008jmlr}. 
Each point in the embedding is assigned its corresponding action label. As shown in Fig.~\ref{fig:method}(B), we define four action labels, 0, 1, 2 and 3, corresponding to action ranges \([ -180^\circ, -135^\circ ) \cup [135^\circ, 180^\circ )\), \([ -135^\circ, -45^\circ )\), \([ -45^\circ, 45^\circ )\) and \([45^\circ, 135^\circ )\), respectively. The t-SNE projection reveals clear grouping by action label, indicating that observations associated with similar actions exhibit similar learned representations. This pattern is consistent with the representation capturing action-relevant information, such as the relative goal direction and local environmental geometry. 
Additional t-SNE results are provided in Supplementary Fig.~S6.

To further evaluate the generalization to unseen scenarios, we conduct experiments in vascular channels extracted from a publicly available retinal fundus image database~\cite{dataset_eye_2017}. 
In Fig.~\ref{fig:method}(C), the two images show the original fundus images and the corresponding extracted vascular regions, respectively. We select five distinct scenarios and sample two trajectories for each. As shown in Fig.~\ref{fig:method}(D), the trained policy successfully navigates the microrobot from its initial position to the designated target in all trials. 
Representative navigation processes are provided in Supplementary Movie~S2.

\subsection{Zero-shot Sim-to-real Transfer}

\begin{figure}[htbp]
    \centering
    \includegraphics[width=\textwidth, trim = 0 0 0 0, clip]{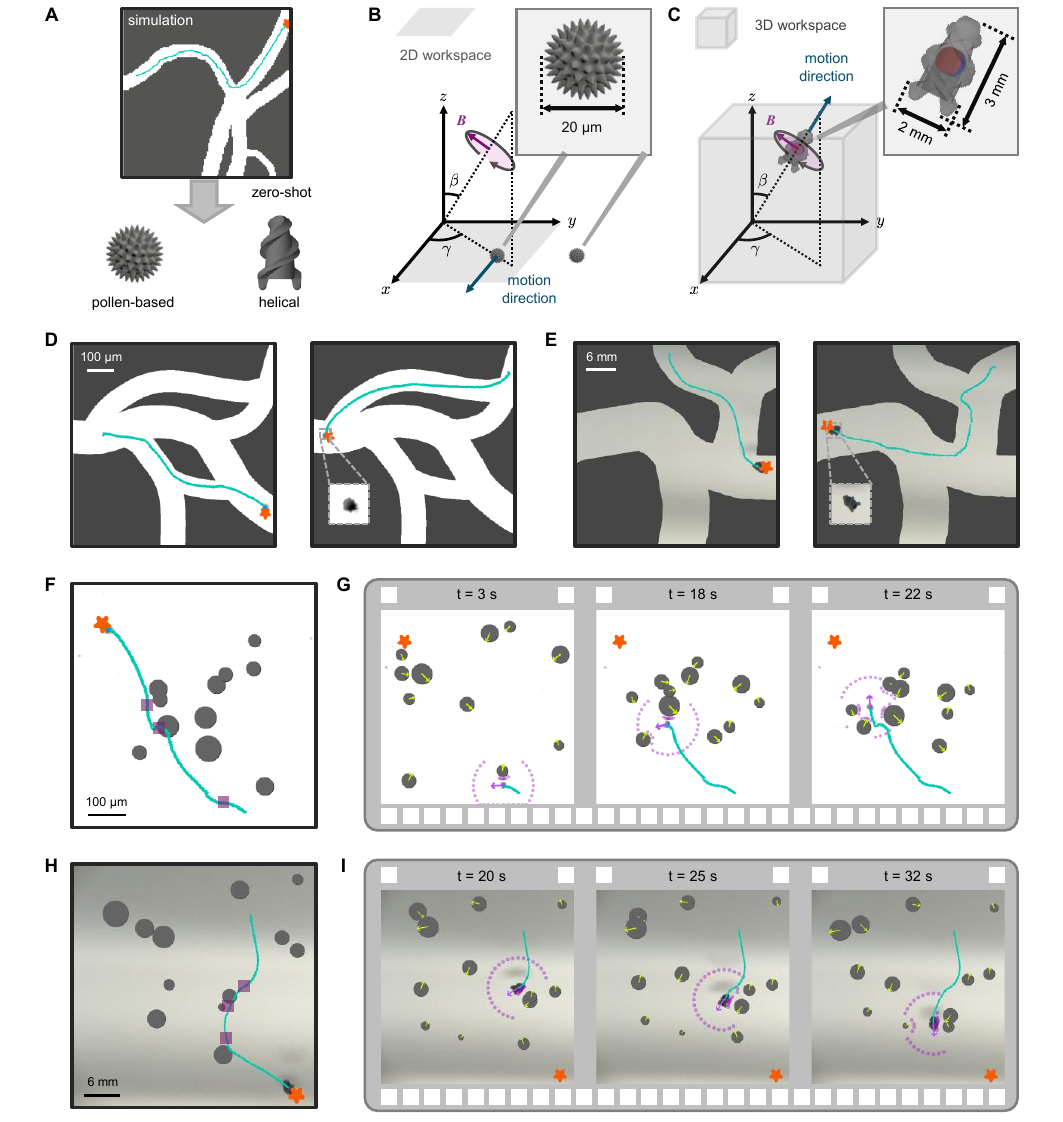}
    \caption{\textbf{Sim-to-real validation across diverse robots and scenarios.} (A) Zero-shot transfer of the trained policy to dissimilar platforms; validation is performed on a pollen-based microparticle and a helical microrobot. (B) Schematic of pollen-based microparticle locomotion under a rotating magnetic field. (C) Schematic of helical microrobot locomotion under a rotating magnetic field. (D) Navigation of the pollen-based microparticle in an artificial vascular channel; the red star marks the target. (E) Navigation of the helical microrobot in an artificial vascular channel. (F) Navigation of the pollen-based microparticle through dynamic obstacles; purple squares denote key frames along the trajectory. (G) Key frames for collision avoidance; yellow arrows indicate obstacle motion, and the purple arrow indicates the action generated by the policy network. (H) Navigation of the helical microrobot through dynamic obstacles. (I) Key frames for collision avoidance.}
    \label{fig:2d_nav}
\end{figure}

\begin{figure}[htbp]
    \centering
    \includegraphics[width=\textwidth, trim = 0 0 0 0, clip]{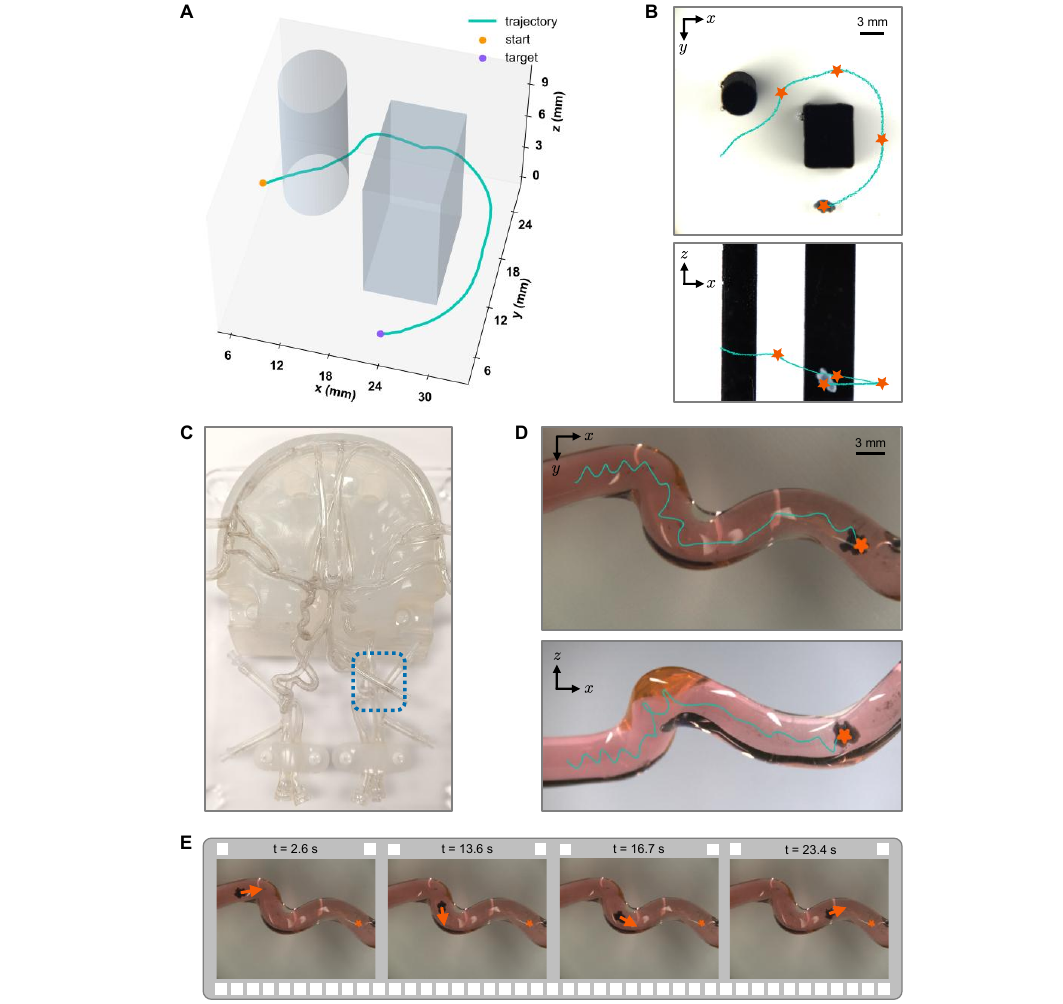}
    \caption{\textbf{Navigation in 3D scenarios.} (A) Visualization of the 3D trajectory in an environment containing two 3D-printed obstacles. (B) Detailed illustration of both the top and side views. (C) Visualization of the human brain vascular model and the specific region (highlighted in a blue square) used in the experiment. (D) Trajectory within a vascular phantom, shown from the top and side views. (E) Snapshots showing the top view of the navigation process within the brain vascular model. The orange arrows denote the actions produced by the policy network.}
    \label{fig:3d_nav}
\end{figure}

To validate whether the learned policy generalizes across microrobot embodiments and working scenarios, we perform zero-shot sim-to-real transfer on two magnetically actuated microrobots and test them in multiple navigation tasks (Fig.~\ref{fig:2d_nav}(A)). 
The first is a pollen-based microparticle (Fig.~\ref{fig:2d_nav}(B)), a \(\text{Fe}_3\text{O}_4\)-coated wild chrysanthemum pollen grain with a diameter of approximately \(20~\mu\text{m}\), which tumbles in the \(x\)-\(y\) plane under a rotating magnetic field. 
The second is a helical microrobot~\cite{aoji_helical_robot_2025npjrobotics} (Fig.~\ref{fig:2d_nav}(C)), which can perform \(3\text{D}\) propulsion along its rotating axis; in this section, we consider its motion projected onto the \(x\)-\(y\) plane. 
For both microrobots, heading is controlled by the yaw angle \(\gamma\), whereas forward speed depends primarily on the rotating field's frequency and pitch angle \(\beta\). 
A \(3\text{D}\) Helmholtz coil system generates the rotating magnetic fields, and closed-loop control is executed at an average rate of \(45~\text{Hz}\).

We first test both microrobots in artificial vascular networks (Fig.~\ref{fig:2d_nav}(D)(E) and Supplementary Movie~S3). 
For each microrobot, we record two trajectories in one previously unseen vascular environment. 
To account for differences in robot size, occupied regions in the map are dilated before generating virtual-LiDAR observations for the policy.
Without any task-specific adaptation, the policy successfully navigates both microrobots, demonstrating zero-shot transfer across microrobot morphologies and environments. 
Long-horizon real-world experiments further show high trajectory consistency across repeated trials, with only minor trial-to-trial variation (Extended Data Fig.~\ref{fig:quant_real} and Supplementary Movie~S8).

We next examine robustness to physical disturbance in a flowing fluid. The helical microrobot is required to traverse a \(5~\text{mm}\)-diameter channel, only \(2.5\) times its body diameter, against a flow of \(2~\text{mm/s}\). 
Although the imposed flow reduces translational speed and destabilizes motion relative to the static-fluid condition, the policy remains effective (Supplementary Fig.~S9 and Supplementary Movie~S7), indicating the robustness to flow-induced perturbations during real-world operation.

Finally, we challenge the policy with dynamic obstacles, a setting not encountered during training. 
Despite this distribution shift, the policy generalizes zero-shot to environments containing \(11\) moving obstacles with random initial positions and velocities (Fig.~\ref{fig:2d_nav}(F)--(I), Supplementary Movie~S4). 
Both the pollen-based and helical microrobots execute online avoidance maneuvers while maintaining progress towards the target. 
This capability arises from the real-time perception pipeline, which converts overhead images into virtual-LiDAR observations that encode local geometric features together with robot and target states, allowing the policy to respond immediately to obstacle motion.
Additional examples are provided in Supplementary Fig.~S4. 
Collectively, these results show that the policy transfers across distinct microrobot platforms and remains effective in previously unseen, dynamic, and physically perturbed environments.

\subsection{Navigation in 3D Environments}
To further assess the effectiveness and real-world performance of the trained policy, we test a helical microrobot in two 3D environments. 
In the first scenario that contains cylindrical and cubic obstacles (Fig.~\ref{fig:3d_nav}(A)), two cameras provide top and side views. Because obstacle-induced occlusion frequently makes side-view height estimation unreliable, control relies on top-view feedback for planar navigation, whereas height follows a pre-defined time-varying command. The microrobot successfully traverses the environment while maintaining clearance from both obstacles (Fig.~\ref{fig:3d_nav}(B)). 
The full experiment is recorded in Supplementary Movie~S5, and details of the image-processing pipeline are provided in Supplementary Text~6 and Supplementary Fig. S2.

Next, we evaluate the policy in a more challenging setting: an anatomically scaled, tortuous cerebral-artery phantom (Fig.~\ref{fig:3d_nav}(C)). 
The confined channel, with a diameter of only approximately \(1.5\times\) the robot length, increases susceptibility to fluid disturbances and makes control more complicated. 
In this experiment, we implement fully autonomous 3D navigation: the trained policy controls the robot's \(x\) and \(y\) position using top-view feedback, and a proportional-derivative (PD) controller (Supplementary Text~7) regulates height from the side view. 
Starting from one end of the vascular channel, the microrobot autonomously reaches a target at the other end (Fig.~\ref{fig:3d_nav}(D); Supplementary Movie~S6). 
The trajectory is less smooth, likely because the relatively large robot size limited manoeuvrability in the narrow phantom. This performance limitation can be mitigated by designing and fabricating smaller-scale robots for improved maneuverability in confined spaces.

\section{Discussion}

Our results demonstrate that a fully vectorized simulator, built on a large-scale vascular dataset, can achieve minute-scale training for microrobot navigation.
Furthermore, we have shown that a systematic study of reward design is crucial for both accelerating training and enhancing the learned policy. Compared with reward formulations commonly used in the existing literature, the proposed TSR reward framework yields substantial improvements in convergence speed, average success rate, and trajectory quality.
Collectively, our framework enables zero-shot deployment across diverse microrobots and scenarios. 

Despite these advances, several limitations highlight important directions for future work. 
First, our current vascular dataset, although large, is artificially constructed based on human vascular features. 
While our experiments indicate that training with this dataset, together with the proposed LiDAR model, can effectively capture local geometric features around the microrobot, a large-scale dataset obtained from real human vascular anatomies would be more beneficial.
In particular, 3D reconstructions of real vasculature would better support fully 3D navigation, and curated datasets from different regions within the same organ relevant to microrobot navigation would allow policies to learn more clinically meaningful structures for region-specific interventions.
Second, within the proposed TSR reward framework, we only present one representative instantiation. 
For different microrobot designs and application scenarios, researchers may tailor alternative regularization terms to improve trajectory quality further and better reflect task-specific constraints or objectives. 
Moreover, although we have demonstrated the strong adaptability of the trained policy, deploying microrobots in real vascular environments remains challenging due to their highly dynamic and uncertain nature. In future work, we will develop more accurate physical models of in vivo vascular conditions and design algorithms for real-time identification of key physical parameters to facilitate better clinical translation. Nevertheless, the proposed minute-scale training framework provides a strong foundation for accelerating algorithmic advances needed to address these challenges. 

\section{Methods}

\subsection{The Virtual LiDAR Model}
The presented virtual LiDAR model simulates ray-casting-based depth sensing. Within each environment, the model emits $N_d$ rays uniformly distributed over $2 \pi$. Each ray originates from the robot's position $\bm{p}_t$ and propagates in the direction $\bm{d}_k = (\cos \theta_k, \sin \theta_k)$, with angles defined by $\theta_k = 2\pi k / N_d$ for $k = 0, 1, \dots, N_d-1$. The ray-casting process iteratively advances each ray in fixed discrete steps until it reaches the maximum detection range $D_{\max}$, collides with an obstacle or non-traversable region, or intersects the boundary of the occupancy map.

To deploy this virtual LiDAR model in real-world experiments, we first convert the raw RGB camera feed into an occupancy map. We then apply the procedure described above to generate the corresponding LiDAR scan points.

\subsection{Simulation Formulation and Implementation}

Modern deep learning frameworks such as PyTorch~\cite{pytorch} and TensorFlow~\cite{tensorflow} provide highly optimized tensor operations backed by C++ kernels and GPU parallelization. We therefore implemented the simulator entirely with tensor-based operations. This design enables the parallel simulation of thousands of microrobots across thousands of vascular environments, yielding large numbers of transitions at each timestep. The complete simulation pipeline is summarized in Algorithm~1. 

Our simulator operates as an iterative cycle with four modules: 
(i) a dynamics update module that computes the microrobot's next state using a unified formulation of its dynamics in the low-Reynolds-number regime:
\begin{equation}
    \begin{aligned}
        v_x &= \mu f \cos \gamma + \xi_x \\
        v_y &= \mu f \sin \gamma + \xi_y ,
    \end{aligned}
    \label{eq:dynamics}
\end{equation}
where $\mu$ is a system-specific constant that can be derived or measured.
$f$ and $\gamma$ are field parameters related to microrobot motion speed (e.g., frequency), and motion direction (e.g., yaw angle).
$\xi_x$ and $\xi_y$ represent the unmodeled dynamics and external disturbances in $x$ and $y$ directions, respectively.
In this work, they are modeled as zero-mean Gaussian random variables, $\xi_x \sim \mathcal{N}(0,\sigma_x^2)$ and $\xi_y \sim \mathcal{N}(0,\sigma_y^2)$. 
The robot position is subsequently updated according to the selected numerical integration scheme.
This model can be applied to various types of microrobots, such as helical \cite{zhao2024automatic, wang2024deep}, rolling \cite{bozuyuk2022reduced, park20253d}, and swarms \cite{yu2018ultra, yu2018pattern}, ensuring the generalization ability of our framework.
(ii) a feasibility-check module that verifies whether the updated state remains within the vascular lumen and reinitializes invalid environments from randomly sampled feasible states; (iii) a visual feature extraction module that processes virtual LiDAR measurements to encode local vascular geometry; and (iv) an observation--reward module that constructs the observation vector and computes the reward under the proposed TSR framework. Owing to full vectorization, the simulator processes on the order of $10^5$ transitions per second, reducing end-to-end training time from hours to minutes. 

In the following, we specifically elaborate on the vectorized implementations for dynamics computation and visual feature extraction, as these two components are the most challenging. 

\textbf{Parallel Dynamics Computation:}
For $n$ microrobots, Equation~(\ref{eq:dynamics}) can be evaluated in batched form as
\begin{equation}
    \begin{bmatrix}
        v_{x_1} \\ v_{x_2} \\ \vdots \\ v_{x_n}
    \end{bmatrix} = \mu f \begin{bmatrix}
        \cos \gamma_1 \\ \cos \gamma_2 \\ \vdots \\ \cos \gamma_n
    \end{bmatrix} + \begin{bmatrix}
        \xi_{x_1} \\ \xi_{x_2} \\ \vdots \\ \xi_{x_n}
    \end{bmatrix}
\end{equation}
where $v_{x_i}$, $\gamma_i$, and $\xi_{{x}_i}$ represent the velocity in $x$-direction, yaw angle, and unmodeled dynamics in $x$-direction of the $i$-th robot, for $i \in \{1, \cdots, n\}$, and $n$ is the total number of robots. For simplicity, we can rewrite the above equation as
\begin{equation}
    \bm{V}_x = \mu f \bm{C}(\bm{\gamma}) + \bm{\Xi}_x
\end{equation}
where $\bm{V}_x = [v_{x_1}, \cdots, v_{x_{n}}] \in \mathbb{R}^n$, $\Xi_x = [\xi_{x_1}, \cdots, \xi_{x_n}] \in \mathbb{R}^n$, $\bm{\gamma} = [\gamma_1, \cdots, \gamma_n] \in \mathbb{R}^n$, and $\bm{C}(\bm{\gamma}) = [\cos\gamma_1, \cdots, \cos\gamma_n] \in \mathbb{R}^n$. Similarly, we can write the matrix form in the $y$-direction:
\begin{equation}
    \bm{V}_y = \mu f \bm{S}(\bm{\gamma}) + \bm{\Xi}_y
\end{equation}
where $\bm{S}(\bm{\gamma}) = [\sin\gamma_1, \cdots, \sin\gamma_n] \in \mathbb{R}^n$, $\bm{V}_y \in \mathbb{R}^n$, and $\Xi_{y} \in \mathbb{R}^n$. In our simulation framework, we only need to compute the four matrices $\bm{S}(\bm{\gamma})$, $\Xi_{x}$, $\bm{C}(\bm{\gamma})$, and $\bm{\Xi}_y$ at each iteration for dynamics updates for thousands of robots simultaneously. It is also notable that the proposed parallel dynamics computation method can be easily generalized to cases where frequency $f$ is also a control variable, as well as to 3D cases. See Supplementary Texts 4 and 5 for extensions to variable-frequency and three-dimensional control.

\textbf{Parallel visual feature extraction.}
To compute visual features across $n$ environments and $N_d$ LiDAR directions simultaneously, we first broadcast the robot position matrix $\bm{P}_t \in \mathbb{R}^{n \times 2}$ into a three-dimensional tensor $\underline{\bm{P}}_t \in \mathbb{R}^{n \times N_d \times 2}$, where each slice along the direction dimension is equal to $\bm{P}_t$. Similarly, we broadcast the LiDAR direction matrix $\bm{D} = [\bm{d}_1, \dots, \bm{d}_{N_d}]^\top \in \mathbb{R}^{N_d \times 2}$ into $\underline{\bm{D}} \in \mathbb{R}^{n \times N_d \times 2}$, where each slice along the environment dimension is equal to $\bm{D}$. Ray casting is then implemented iteratively as
\begin{equation}
    \begin{aligned}
        \bm{M}_{t,1} &= \underline{\bm{P}}_t, \\
        \bm{M}_{t,m+1} &= \bm{M}_{t,m} + \mathbb{I}_{\mathcal{V}}\left\{\left\lfloor \bm{M}_{t,m} + \underline{\bm{D}} \right\rceil \in \mathcal{V}\right\} \odot \underline{\bm{D}},
    \end{aligned}
\end{equation}
where $\bm{M}_{t,m} \in \mathbb{R}^{n \times N_d \times 2}$ denotes the tensor of LiDAR sample points for all environments and all directions at time $t$ and iteration step $m$; $\mathcal{V}$ denotes the feasible region for microrobot motion; $\lfloor \cdot \rceil$ is the rounding operator; $\odot$ denotes elementwise multiplication; and $\mathbb{I}_{\mathcal{V}}\{\cdot\}$ is an indicator function that returns a binary mask, with value 1 when the rounded position remains within $\mathcal{V}$ and 0 otherwise. This masking operation prevents rays from propagating beyond the feasible region while preserving full parallelism across environments and directions.

\begin{algorithm}[tbp]
\caption{Vectorized simulation for microrobot navigation}
\label{alg:sim}
\small
\begin{algorithmic}[1]
\Require action $\bm{a}_t = \bm{\gamma}_t \in \mathbb{R}^{n}$
\Ensure states $\bm{s}_{t+1}$, rewards $r_{t+1}$, and logging information

\AlgPhase{algblue}{algblueback}{Initialization}
\State \textbf{if} this is the first simulation step \textbf{then}
\State \hspace*{1.25em} Import the vascular dataset and allocate it to the parallel environments
\State \hspace*{1.25em} Extract the feasible vascular regions $\mathcal{V} = \{\mathcal{V}_1, \cdots, \mathcal{V}_n\}$
\State \hspace*{1.25em} $\mathcal{I}_{\mathrm{reset}} \gets \{1, 2, \ldots, n\}$
\State \hspace*{1.25em} $\bm{d}_k \gets (\cos \theta_k, \sin \theta_k)$, where $\theta_k = 2\pi k/N_d$
\State \hspace*{1.25em} $\bm{D} \gets [\bm{d}_1, \ldots, \bm{d}_{N_d}] \in \mathbb{R}^{N_d \times 2}$
\State \hspace*{1.25em} Broadcast $\bm{D}$ into $\underline{\bm{D}} \in \mathbb{R}^{n \times N_d \times 2}$
\State \hspace*{1.25em} Define the maximum episode length $L_{\max}$
\State \hspace*{1.25em} \Call{Reset}{$\mathcal{I}_{\mathrm{reset}}$}
\State \textbf{end if}


\AlgPhase{algteal}{algtealback}{Vectorized dynamics update}
\State Sample $\bm{\Xi}_x$ and $\bm{\Xi}_y$ from their respective distributions
\State $\bm{V}_x \gets \mu f\bm{C}(\bm{\gamma}_t) + \bm{\Xi}_x$
\State $\bm{V}_y \gets \mu f\bm{S}(\bm{\gamma}_t) + \bm{\Xi}_y$
\State $\bm{V}_{\mathrm{robot},t} \gets [\bm{V}_x,\bm{V}_y]$
\State $\bm{P}_{\mathrm{robot},t+1} \gets \bm{P}_{\mathrm{robot},t}
    + \bm{V}_{\mathrm{robot},t}\Delta t$    

\AlgPhase{alggold}{alggoldback}{Vectorized LiDAR perception}
\State Broadcast $\bm{P}_{\mathrm{robot},t+1}$ into
    $\underline{\bm{P}}_{t+1} \in \mathbb{R}^{n \times N_d \times 2}$
\State $\bm{M}_{t+1,0} \gets \underline{\bm{P}}_{t+1}$
\State \textbf{for} $m \gets 0$ \textbf{to} $D_{\max}-1$ \textbf{do}
\State \hspace*{1.25em} $\bm{M}_{t+1,m+1} \gets \bm{M}_{t+1,m}
        + \mathbb{I}_{\mathcal{V}}\!\left\{\left\lfloor\bm{M}_{t+1,m}
        + \underline{\bm{D}}\right\rceil \in \mathcal{V}\right\}
        \odot \underline{\bm{D}}$
\State \textbf{end for}

\AlgPhase{algslate}{algslateback}{Termination and output}
\State $l_i \gets l_i + 1$ for each of the vectorized environments
\State $\mathcal{I}_{\mathrm{collision}} \gets
    \{i \mid \bm{P}_{\mathrm{robot},t+1}[i,:] \notin \mathcal{V}_i\}$
\State $\mathcal{I}_{\mathrm{success}} \gets
    \{i \mid \|\bm{P}_{\mathrm{robot},t+1}[i,:]-\bm{P}_{\mathrm{target}}[i,:]\|_2 \leq \delta\}$
\State $\mathcal{I}_{\mathrm{timeout}} \gets \{i | l_i \geq L_{\max} \}$
\State $\bm{s}_{t+1} \in \mathbb{R}^{n\times|\mathcal{S}|} \gets
    \Call{ComputeStates}{\bm{P}_{\mathrm{robot},t+1},
    \bm{P}_{\mathrm{target}},\bm{M}_{t+1,D_{\max}},\bm{a}_t}$
\State $\bm{r}_{t+1} \in \mathbb{R}^{n} \gets
    \Call{ComputeRewards}{\bm{s}_t, \bm{a}_t, \bm{s}_{t+1}}$
\State $\mathcal{I}_{\mathrm{reset}} \gets
    \mathcal{I}_{\mathrm{collision}} \cup \mathcal{I}_{\mathrm{success}} \cup \mathcal{I}_{\mathrm{timeout}}$
\State \Call{Reset}{$\mathcal{I}_{\mathrm{reset}}$}

\end{algorithmic}
\end{algorithm}

\subsection{RL Formulation}
We formulate the navigation task as a discrete-time problem, where the environment at each timestep $t$ is fully characterized by the state $\bm{s}_t$. At every step, the navigation policy $\pi$ selects an action $\bm{a}_t$, causing the environment to transition to a new state $\bm{s}_{t+1}$ based on the state-transition dynamics $p(\bm{s}_{t+1} \mid \bm{s}_t, \bm{a}_t)$. Concurrently, the environment provides the robot with a feedback reward signal $r_{t+1}$. The objective is to identify an optimal policy $\pi^\star$ that maximizes the expected discounted cumulative reward over future trajectories. The specific components of this formulation are detailed below.

\noindent\textbf{State:} The state $\bm{s}_t$ comprises the robot's current position $\bm{p}_{\text{robot}, t} \in \mathbb{R}^2$, the target position $\bm{p}_\text{target} \in \mathbb{R}^2$, virtual LiDAR scan points $\bm{m}_t \in \mathbb{R}^{{N}_d\times2}$, and the previous action $\bm{a}_{t-1} \in \mathbb{R}$. Setting the number of virtual LiDAR scan points to $N_d=36$ yields a 77-dimensional observation vector at each timestep $t$:
\begin{equation}
    \bm{s}_t = [\bm{p}_{\text{robot},t}, \bm{p}_{\text{target}}, \bm{m}_t, \bm{a}_{t-1}]
\end{equation}

\noindent\textbf{Action:} For microrobot navigation under a rotating magnetic field with constant frequency, the action space is defined by the desired yaw angle of the magnetic field used for actuation. We employ a continuous action space where actions can take any real value within the interval $[-\pi, \pi)$. Unlike many traditional approaches that discretize the action space~\cite{dqn_2d_time_penalty_2023icra, dqn_2d_time_penalty_2024engineering}, utilizing a continuous space expands the search domain and increases problem complexity. However, by implementing efficient training strategies, our method ensures convergence within minutes.

\noindent\textbf{Reward:} We present the detailed TSR reward framework. The task-oriented reward is defined as
\begin{equation}
    r_\text{task}(\bm{s}_t, \bm{a}_t, \bm{s}_{t+1}) = 
    \begin{cases}
        +1 & \text{if} \ \|\bm{p}_{\text{robot}, t+1} - \bm{p}_\text{target}\|_2 \leq \delta \\
        -1 & \text{if} \ \bm{p}_{\text{robot}, t+1} \notin \mathcal{V} \\
        0  & \text{otherwise}
    \end{cases}
\end{equation}
where $\bm{p}_{\text{robot}, t}$ and $\bm{p}_\text{target}$ denote the robot and target positions, respectively, $\delta$ is a predefined threshold controlling the navigation precision, and $\mathcal{V}$ represents the feasible vascular domain. Once the robot exits $\mathcal{V}$, the episode terminates immediately with a penalty, as illustrated in Fig.~\ref{fig:teaser}(C).

The shaping component provides dense transition-level feedback to improve temporal credit assignment. We instantiate this component using a unit-discount potential-based shaping formulation:
\begin{equation}
    r_{\text{shape},t+1}
    =
    \gamma_s \Phi(\bm{s}_{t+1})
    -
    \Phi(\bm{s}_t),
\end{equation}
where $\gamma_s$ denotes the shaping discount factor used exclusively in constructing this reward term and is distinct from the return discount factor used in PPO. We define the potential function as the negative Euclidean distance between the microrobot and the target:
\begin{equation}
    \Phi(\bm{s}_t)
    =
    -\left\|
    \bm{p}_{\text{robot},t}
    -
    \bm{p}_{\text{target}}
    \right\|_2.
\end{equation}
In this work, we set $\gamma_s=1$, such that the shaping reward becomes
\begin{equation}
    r_{\text{shape},t+1}
    =
    d_t-d_{t+1},
\end{equation}
where
$d_t=\|\bm{p}_{\text{robot},t}-\bm{p}_{\text{target}}\|_2$
is the robot--target distance at timestep $t$. The shaping reward is therefore positive when the microrobot moves closer to the target, negative when it moves farther away, and zero when the distance remains unchanged. Under this unit-discount setting, the shaping signal exactly measures the geometric progress between consecutive states, providing a clear physical interpretation for microrobot navigation. As demonstrated in our experiments, this formulation enables faster and more stable learning.

In this work, we consider two regularization terms: an action-smoothness reward and an obstacle-safety reward. The action-smoothness reward penalizes large deviations between successive actions:
\begin{equation}
    r_\text{smooth} = -\| \bm{a}_t - \bm{a}_{t-1} \|_2
\end{equation}
The obstacle-safety reward encourages the agent to keep away from nearby structures:
\begin{equation}
    r_\text{safety} = \frac{1}{N_d} \sum_{i=1}^{N_d} \log \| \bm{m}_i - \bm{p}_{\text{robot}, t} \|_2
\label{eq:rew_safety}
\end{equation}
where $\bm{m}_i$ denotes LiDAR scan points, and $N_d$ is the total number of scan points. By averaging the logarithmic distances between the agent and each scan point, this term promotes trajectories that maintain a safety buffer from potential hazards.

\noindent\textbf{State-transition dynamics:} The state-transition dynamics follow the motion characteristics of the microrobot. In discrete time, the robot's position is updated as:
\begin{equation}
    \bm{p}_{\text{robot}, t+1} = \bm{p}_{\text{robot}, t} + \bm{v}_t \, \text{d}t
\end{equation}
where $\bm{v}_t = [v_x, v_y]^T$ denotes the robot’s velocity computed from Equation~(\ref{eq:dynamics}), and $\text{d}t$ represents the control period. The LiDAR model is then applied to update the scan points $\bm{m}_t$. The target position $\bm{p}_{\text{target}}$ remains unchanged unless the episode terminates, while the previous action $\bm{a}_{t-1}$ is obtained from the output of the policy network.

\noindent\textbf{Episode Termination:} An episode terminates if one of three conditions is met: the robot reaches the goal, a collision occurs, or the maximum trajectory length is exceeded. Specifically, the goal is considered reached when the robot's position falls within a predefined distance threshold from the target, which establishes the required navigation precision. The maximum trajectory length constraint prevents indefinite execution, thereby ensuring computational efficiency.

\subsection{Policy Optimization}
The control policy is learned via the PPO~\cite{ppo_arxiv2017} algorithm. The framework utilizes an actor-critic architecture, where both the policy network $\pi_{\bm{\theta}}$ (actor) and the value network $V_{\bm{w}}$ (critic) are parameterized as multi-layer perceptrons (MLP) taking observations $\bm{s}_t$ as input. To guide policy updates, we compute the advantage estimate $\hat{A}_t$, which quantifies the excess return of executing action $\bm{a}_t$ relative to the state's baseline value $V_{\bm{w}}(\bm{s}_t)$. Data collection is parallelized across $n = 4,096$ environments. In each iteration, we collect trajectories of $T=48$ timesteps to compile a batch of 196,608 transitions. The actor minimizes the negative clipped surrogate objective:
\begin{equation}
\mathcal{L}_\text{actor}(\bm{\theta}) = - \mathbb{E}_t \left[ \min \left( \rho_t(\bm{\theta}) \hat{A}_t, \text{clip}\left( \rho_t(\bm{\theta}), 1 - \epsilon, 1 + \epsilon \right) \hat{A}_t \right) \right],
\label{eq:loss_actor}
\end{equation}
where $\rho_t(\bm{\theta}) = \frac{\pi_{\bm{\theta}}(\bm{a}_t|\bm{s}_t)}{\pi_{\bm{\theta}_{\text{old}}}(\bm{a}_t|\bm{s}_t)}$ denotes the probability ratio between the new and old policies, the $\text{clip}(\cdot)$ operator controls the extent of the policy update, and $\epsilon$ is the clipping hyperparameter. The critic minimizes the mean squared error between the estimated value and the target returns $R_t$:
\begin{equation}
\mathcal{L}_\text{critic}(\bm{w}) = \mathbb{E}_t \left[ \left( V_{\bm{w}}(\bm{s}_t) - R_t \right)^2 \right].
\label{eq:loss_critic}
\end{equation}
We also leverage an entropy bonus to encourage exploration. 
In each training iteration, 48 transitions are collected from each of 4,096 environments to form an on-policy batch of 196,608 transitions, after which the actor-critic is optimized on this batch for 5 epochs using mini-batches of 49,152 transitions.
Training was performed on a system equipped with an Intel(R) Core(TM) i7-14700 CPU, 32 GB RAM, and a single NVIDIA GeForce RTX 4060 GPU.
Detailed hyperparameter settings and network architectures are provided in Supplementary Tables~S1 and S2, respectively.

\subsection{Magnetic Actuation Framework}
The locomotion of the microrobot is governed by the spatial orientation of an external rotating magnetic field $\bm{B}$. To achieve omnidirectional navigation, we synthesize an arbitrary 3D rotating field by superimposing the orthogonal magnetic components generated by a tri-axial Helmholtz coil system.

We first define a planar base field, $\bm{B}_{\text{base}}$, rotating within the $xy$-plane at a frequency $f$:
\begin{equation}
\bm{B}_{\text{base}} = [B_0 \cos(2\pi f t), \ B_0 \sin(2\pi f t), \ 0]^T
\end{equation}
where $B_0$ denotes the magnetic field amplitude. The desired 3D orientation is achieved by applying a sequential rotation transformation to $\bm{B}_{\text{base}}$: a pitch rotation $\beta$ about the $y$-axis, followed by a yaw rotation $\gamma$ about the $z$-axis. This transformation is expressed as:
\begin{equation}
\bm{B} = \bm{R}_z(\gamma) \bm{R}_y(\beta) \bm{B}_{\text{base}} =
\begin{bmatrix}
\cos \gamma & -\sin \gamma & 0 \\
\sin \gamma & \cos \gamma & 0 \\
0 & 0 & 1
\end{bmatrix}
\begin{bmatrix}
\cos \beta & 0 & \sin \beta \\
0 & 1 & 0 \\
-\sin \beta & 0 & \cos \beta
\end{bmatrix} \bm{B}_{\text{base}}
\end{equation}
Expanding this matrix product yields the time-varying magnetic field components required for the $x$, $y$, and $z$ axes:
\begin{equation}
\bm{B} = 
\begin{bmatrix}
B_x(t) \\ B_y(t) \\ B_z(t)
\end{bmatrix} =
B_0 \begin{bmatrix}
\cos \gamma \cos \beta \cos(2\pi f t) - \sin\gamma\sin(2\pi ft) \\
\sin \gamma \cos \beta \cos(2\pi f t) + \cos\gamma\sin(2\pi ft) \\
-\sin\beta \cos(2\pi f t)
\end{bmatrix}
\end{equation}
These components ($B_x, B_y, B_z$) are synchronized to actuate the microrobot along the prescribed heading vector (as illustrated in Fig.~\ref{fig:2d_nav}(B)(C)).

\begin{extfigure}[htbp]
    \centering
    \includegraphics[width=\textwidth, trim = 0 0 0 0, clip]{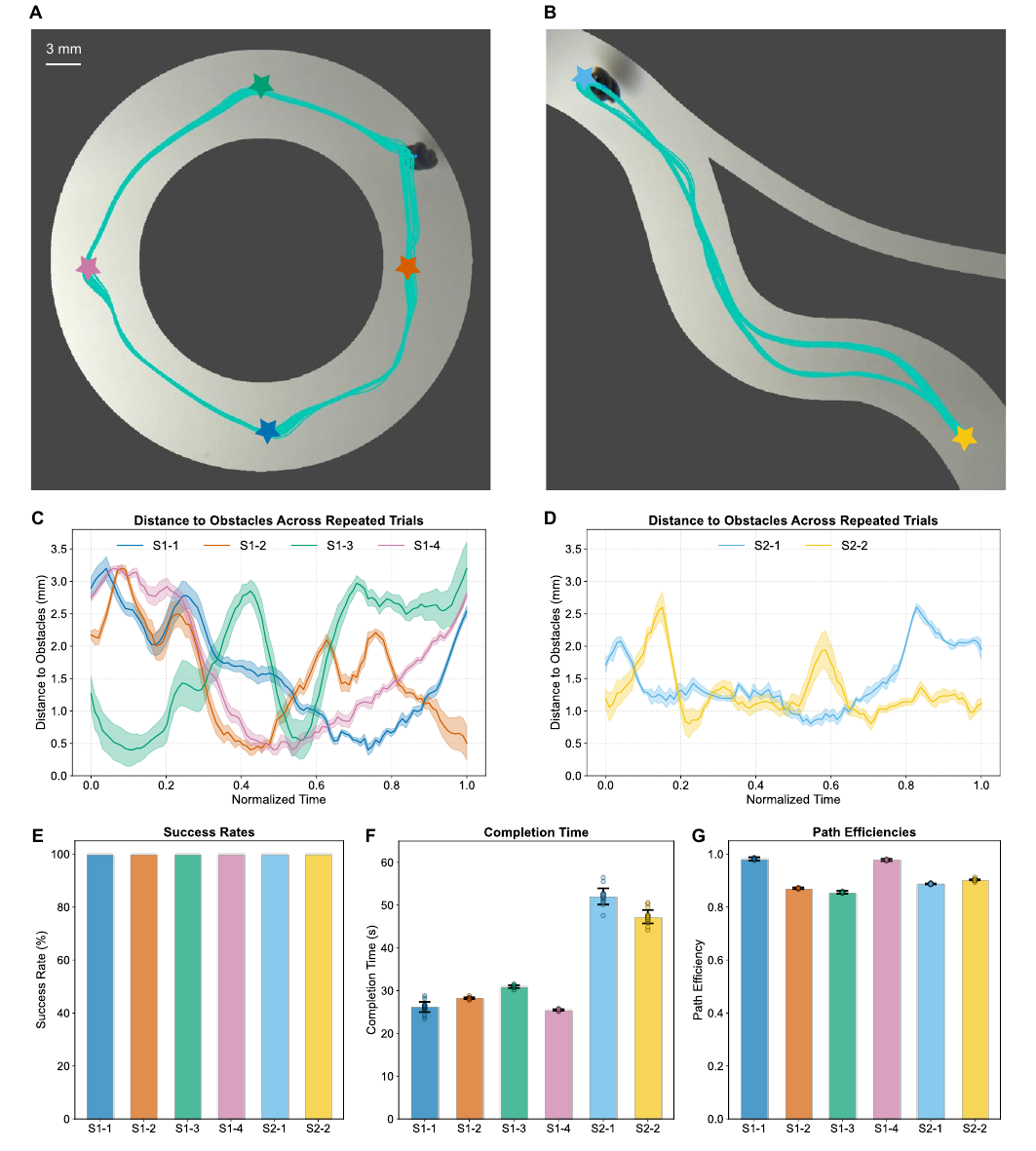}
    \caption{
    \textbf{Repeated-trial analysis of long-horizon navigation.} S$i$-$j$ in legends represents the $j$-th trajectory in the scenario $i$. We collect 15 trials for each trajectory and analyze the trial-to-trial variability and performance metrics. Please see the Supplementary Movie S8. (A)-(B) Representative navigation trajectories in a circular environment and a channel environment. (C)-(D) Obstacle clearance over normalized time during repeated executions of the trajectories in the circular (C) and channel (D) environments. Each colour represents one trajectory, as labelled above the plots. Solid lines show the mean across 15 independent trials, and shaded regions indicate the corresponding 95\% confidence intervals. (E)-(G) Quantitative evaluation of real-world navigation performance: (E) success rate, (F) completion time and (G) path efficiency. Bars and error bars indicate the mean ± s.d. across 15 independent trials, respectively, and dots represent individual trial values.}
    \label{fig:quant_real}
\end{extfigure}

\clearpage

\noindent\textbf{Data Availability}\\
All data generated during the study are available in the main text, the Supplementary Information and the GitHub repository \href{https://github.com/yinghansun/mr-nav}{https://github.com/yinghansun/mr-nav}. 
The large-scale vascular dataset for policy training is available from Zenodo at \href{https://doi.org/10.5281/zenodo.21719761}{https://doi.org/10.5281/zenodo.21719761}~\cite{sun2026large}.
Reward function definitions and training hyperparameters are provided in the Supplementary Information. \\

\noindent\textbf{Code Availability}\\
Codes are available on GitHub: \href{https://github.com/yinghansun/mr-nav}{https://github.com/yinghansun/mr-nav}.\\

\noindent\textbf{Acknowledgments}\\
We thank the PolyU Industrial Centre for its support.\\

\noindent\textbf{Funding Statement}\\
This work is financially supported by the Research and Innovation Office (Project code: BBRF and BBX2) of The Hong Kong Polytechnic University, the Research Institute for Advanced Manufacturing of Hong Kong Polytechnic University (grant no. 1-CDK3), the Guangdong Basic and Applied Basic Research Foundation (grant no. 2025A1515010116 and 2023A1515110709), and the Hong Kong Research Grant Council (grant nos. STG1/E-401/23-N, 25200424, 15206223, and C4038-24G).\\

\noindent\textbf{Author contributions}\\
Y.S. and L.Y. conceived the study. 
Y.S., A.Z., X.J., Y.L, and J.Z. designed and implemented the system hardware and software, and performed the experiments. 
A.Z. and Y.W. fabricated the microrobots. 
Y.S. designed and coded the DRL algorithms. 
Y.S. wrote the manuscript with contributions from all authors. 
L.Z., H.G., and L.Y. revised the manuscript and supervised the project. All authors contributed to the scientific discussion.\\

\noindent\textbf{Competing interests}\\
There are no competing interests to declare. \\


\bibliography{ref} 

\clearpage


\newpage


\clearpage
\thispagestyle{empty}
\begingroup
\raggedright
\setlength{\parindent}{0pt}
\sffamily

{\fontsize{22}{26}\selectfont\bfseries Supplementary Information for\par}
\vspace{0.55cm}
{\Large\bfseries Minute-Scale Training for Microrobot Navigation\par}

\vspace{0.45cm}
{\small\bfseries
Yinghan~Sun$^{1}$, Aoji~Zhu$^{1}$, Xiang~Ji$^{2}$, Yamei~Li$^{1}$,
Jiachi~Zhao$^{1}$, Yun~Wang$^{1}$, Li~Zhang$^{2,3,*}$,
Huijun~Gao$^{4,*}$, Lidong~Yang$^{1,5,6,*}$\par
}

\vspace{0.35cm}
{\footnotesize
$^{1}$State Key Laboratory of Ultra-precision Machining Technology, Department of Industrial and Systems Engineering, The Hong Kong Polytechnic University, Hong Kong, China.\par
$^{2}$Department of Mechanical and Automation Engineering, The Chinese University of Hong Kong, Hong Kong, China.\par
$^{3}$Shenzhen Loop Area Institute, Shenzhen, China.\par
$^{4}$Research Institute of Intelligent Control and Systems, Harbin Institute of Technology, Harbin, China.\par
$^{5}$Research Institute for Advanced Manufacturing, The Hong Kong Polytechnic University, Hong Kong, China.\par
$^{6}$PolyU-PUTH Medicine-Engineering Collaborative Innovation Research Laboratory, The Hong Kong Polytechnic University, Hong Kong, China.\par
}

\vspace{0.35cm}
{\footnotesize\bfseries
$^{*}$Corresponding authors. Email: lizhang@cuhk.edu.hk; hjgao@hit.edu.cn; lidong.yang@polyu.edu.hk\par
}

\vspace{1.15cm}
{\small\bfseries This PDF file includes:\par}
\vspace{0.15cm}
{\small\rmfamily
\hspace*{0.7cm}Supplementary Text\par
\hspace*{0.7cm}Figures S1 to S9\par
\hspace*{0.7cm}Tables S1 to S4\par
\hspace*{0.7cm}Captions for Movies S1 to S8\par
}

\vspace{0.55cm}
{\small\bfseries Other supporting materials for this manuscript include the following:\par}
\vspace{0.15cm}
{\small\rmfamily
\hspace*{0.7cm}Movies S1 to S8\par
}

\vfill
\hfill{\small\bfseries \thepage\ of \pageref*{LastPage}}\par
\vspace*{0.2cm}
\endgroup
\clearpage

\renewcommand{\thefigure}{S\arabic{figure}}
\renewcommand{\thetable}{S\arabic{table}}
\renewcommand{\theequation}{S\arabic{equation}}
\renewcommand{\theHsection}{supplement.\arabic{section}}
\renewcommand{\theHsubsection}{supplement.\arabic{section}.\arabic{subsection}}
\renewcommand{\theHsubsubsection}{supplement.\arabic{section}.\arabic{subsection}.\arabic{subsubsection}}
\renewcommand{\theHfigure}{supplement.\arabic{figure}}
\renewcommand{\theHtable}{supplement.\arabic{table}}
\renewcommand{\theHequation}{supplement.\arabic{equation}}
\setcounter{section}{0}
\setcounter{subsection}{0}
\setcounter{subsubsection}{0}
\setcounter{figure}{0}
\setcounter{table}{0}
\setcounter{equation}{0}

\section{Experimental Setup}

\textbf{Pollen-based microparticle setup: } The magnetic actuation platform comprises three orthogonal Helmholtz coil pairs for generating approximately uniform magnetic fields to control the pollen-based microparticle. Experiments are performed in an acrylic tank that houses a silicon substrate immersed in a 0.3\% Sodium Dodecyl Sulfate (SDS) solution, where the pollen-based microparticle is actuated. Real-time images of the workspace are captured by an overhead microscope–camera system and streamed to a host computer, which extracts the position of the microrobot for feedback control. The host computer implements the policy inference framework, which converts the camera feed into control signals for magnetic actuation. These signals are sent through an I/O interface card (Model 826, Sensoray, Inc.) to three servo amplifiers (ADS 50/5 4-Q-DC, Maxon, Inc.), which drive the 3D Helmholtz coil system to generate time-varying magnetic fields that actuate the pollen-based microparticle.

\textbf{Helical microrobot setup: }
The effective workspace within the coil assembly is approximately 40~mm $\times$ 40~mm $\times$ 40~mm.
The $x$- and $y$-axis coils are driven by servo current drivers (Advanced Motion Controls, A50A100), while the $z$-axis coils are driven by a lower-current unit (Advanced Motion Controls, A25A100).
Current commands are generated using a data acquisition card (Advantech, PCI-1720U) to provide synchronized control across the three axes.
Imaging feedback is acquired with two industrial cameras (Daheng Imaging, ME2L-161-61U3C-L) fitted with a telecentric lens (Daheng Optics, GCO-232106) to reduce perspective-dependent measurement errors over the workspace.

\clearpage
\section{Fabrication of the Pollen-based Microparticle}
\noindent\textbf{Pollen template purification:}
Chrysanthemum pollen (5.0\,g) is dispersed in anhydrous ethanol (100\,mL) and ultrasonicated for 30\,min.
The suspension is centrifuged (5000\,rpm, 5\,min), the supernatant is discarded, and the pellet is redispersed in deionized water (100\,mL) followed by ultrasonication for 10\,min.
This wash cycle is repeated three times to remove surface impurities, after which the pollen is freeze-dried ($-50\,^{\circ}$C, 12\,h) to yield purified pollen templates.

\noindent\textbf{Chemical bath deposition of Fe$_3$O$_4$:}
Pretreated pollen (200\,mg) is dispersed in deionized water (60\,mL) and ultrasonicated for 10\,min to obtain a homogeneous suspension.
Under continuous stirring, FeSO$_4\cdot$7H$_2$O (183\,mg) is added and reacted for 20\,min to promote adsorption of Fe$^{2+}$ onto the pollen surface.
Aqueous ammonia (25--27\,wt\%, 20\,mL) is then added dropwise over 10\,min, and the sealed vessel is stirred at 250\,rpm for 2\,h to drive in situ formation of Fe$_3$O$_4$ nanoparticles.
The black-brown product is collected with an external magnet and washed alternately with anhydrous ethanol and deionized water (three cycles) before freeze-drying to obtain the magnetic pollen composite.

\clearpage
\section{Fabrication of the Helical Microrobot}
The helical microrobot is fabricated using DLP‑based photopolymerization 3D printing technology with an ELEGOO Mars 4 DLP printer, using an 8K rigid photopolymer resin from the same manufacturer.
To achieve magnetic actuation, the helical microrobot is embedded with N52‑grade neodymium‑iron‑boron (NdFeB) magnets, which provide high energy density and strong magnetic properties.
The final design, determined after numerical optimization of the helical body, yields a robot with an overall size of approximately 2 mm in diameter and 3 mm in length.

\clearpage
\section{Parallel Dynamics Computation for Variable Frequency}
To efficiently address the dynamics computation under varying actuation frequencies, we formulate a vectorized strategy. The following equation governs the velocity components in the $x$-direction for $n$ microrobots:
\begin{equation}
    \begin{bmatrix}
        v_{x_1} \\ v_{x_2} \\ \vdots \\ v_{x_n}
    \end{bmatrix} 
    = \mu 
    \begin{bmatrix}
        f_1 & 0 & \cdots & 0 \\
        0 & f_2 & \cdots & 0 \\
        \vdots & \vdots & \ddots & \vdots \\
        0 & 0 & \cdots & f_n
    \end{bmatrix} 
    \begin{bmatrix}
        \cos \gamma_1 \\ \cos \gamma_2 \\ \vdots \\ \cos \gamma_n
    \end{bmatrix} 
    + 
    \begin{bmatrix}
        \xi_{x_1} \\ \xi_{x_2} \\ \vdots \\ \xi_{x_n}
    \end{bmatrix}
\end{equation}
where $f_i$ denotes the rotational frequency actuating the $i$-th microrobot. This relationship can be expressed in a compact matrix form as:
\begin{equation}
    \bm{V}_x = \mu \bm{F} \bm{C}(\bm{\gamma}) + \bm{\Xi}_x
\end{equation}
where $\bm{F}$ denotes the frequency matrix. Analogously, the dynamics for the $y$-direction are given by:
\begin{equation}
    \bm{V}_y = \mu \bm{F} \bm{S}(\bm{\gamma}) + \bm{\Xi}_y
\end{equation}

\clearpage
\section{The 3D Dynamics Model}
In this paper, the focus is on designing a navigation policy in the $x$-$y$ plane, while the pitch angle is adjusted experimentally to ensure effective microrobot propulsion. When extending this policy to a three-dimensional workspace (Fig.~5(C)), the full 3D dynamics of microrobots under a rotating magnetic field must be taken into account:
\begin{equation}
    \begin{aligned}
        v_x &= \mu f \sin \beta \cos \gamma + \xi_x \\
        v_y &= \mu f \sin \beta \sin \gamma + \xi_y \\
        v_z &= \mu f \cos \beta - v_g + \xi_z
    \end{aligned}
    \label{eq:3d_dynamics}
\end{equation}
where $v_x$, $v_y$, and $v_z$ denote the velocity components along the $x$, $y$, and $z$ axes, respectively, $\mu$ is a system-specific constant that can be identified experimentally, $f$  is the rotation frequency of the magnetic field, $\beta$ and $\gamma$ are the pitch and yaw angles, $v_g$ is the gravity-induced settling velocity, and $\xi_x$, $\xi_y$, and $\xi_z$ represent the unmodeled dynamics and external disturbances in the corresponding directions. 

To derive a compact representation, consider first the dynamics in the $x$-direction for a batch of $n$ microrobots:
\begin{equation}
    \begin{bmatrix}
        v_{x_1} \\ v_{x_2} \\ \vdots \\ v_{x_n}
    \end{bmatrix} 
    = \mu 
    \begin{bmatrix}
        f_1 & 0 & \cdots & 0 \\
        0 & f_2 & \cdots & 0 \\
        \vdots & \vdots & \ddots & \vdots \\
        0 & 0 & \cdots & f_n
    \end{bmatrix}     \begin{bmatrix}
        \sin \beta_1 & 0 & \cdots & 0 \\
        0 & \sin \beta_2 & \cdots & 0 \\
        \vdots & \vdots & \ddots & \vdots \\
        0 & 0 & \cdots & \sin \beta_n
    \end{bmatrix} 
    \begin{bmatrix}
        \cos \gamma_1 \\ \cos \gamma_2 \\ \vdots \\ \cos \gamma_n
    \end{bmatrix} 
    + 
    \begin{bmatrix}
        \xi_{x_1} \\ \xi_{x_2} \\ \vdots \\ \xi_{x_n}
    \end{bmatrix}
\end{equation}
For notational convenience, this expression can be written in vector form as
\begin{equation}
    \bm{V}_x = \mu \bm{F} \text{diag}(\bm{S}(\beta)) \bm{C}(\bm{\gamma}) + \bm{\Xi}_x
\end{equation}
where $\text{diag}(\bm{S}(\beta))$ denotes the diagonal matrix whose diagonal entries are the components of $\bm{S}(\beta)$. 

Analogously, the vectorized dynamics in the $y$-direction are given by
\begin{equation}
    \bm{V}_y = \mu \bm{F} \text{diag}(\bm{S}(\beta)) \bm{S}(\bm{\gamma}) + \bm{\Xi}_y
\end{equation}
Finally, the dynamics along the $z$-direction can be written as
\begin{equation}
    \bm{V}_z = \mu \bm{F} \bm{C}(\beta) - \bm{V}_g + \bm{\Xi}_z
\end{equation}
where $\bm{V}_g = [v_g, v_g, \cdots, v_g]^T \in \mathbb{R}^n$  denotes the gravity-induced settling velocity replicated for all microrobots.

\clearpage
\section{The Image Processing Pipeline}
The robot state and obstacle configuration are inferred directly from monocular overhead images using a deterministic image‑processing pipeline implemented in Python/OpenCV. The overall process is illustrated in Fig.~\ref{fig:img_processing}. Raw color frames are first converted to a single‑channel grayscale image and intensity‑inverted so that the dark robot and obstacles appear bright on a dark background, which facilitates subsequent thresholding and contour detection. To suppress sensor noise while preserving shape boundaries, the inverted image is smoothed with a small Gaussian filter before binarization.

To obtain a robust foreground mask under non‑uniform illumination, the blurred image is binarized using an adaptive Gaussian thresholding scheme, where the threshold at each pixel is set to a Gaussian‑weighted average of intensities in a local neighborhood (here 31×31 pixels) minus a constant offset. This local, data‑driven threshold compensates for slowly varying background intensity and reflections and yields a high‑contrast binary image of the robot and obstacles. The binary mask is then refined by a morphological closing operation with an elliptical structuring element of size 9×9 pixels applied three times, which connects fragmented edges, fills small gaps, and removes spurious holes inside large objects.

Object outlines are extracted from the refined binary mask by tracing all connected foreground boundaries and representing each as a contour curve. The contours are sorted by enclosed area in descending order and classified using empirically chosen area thresholds. This area‑based classification reliably separates the compact robot from the larger square and cylindrical obstacles, as illustrated in Supplementary Fig.~\ref{fig:img_processing}.

For the contour identified as the robot, a minimum‑area rotated rectangle is fitted, corresponding to the smallest possible rectangle that encloses the entire contour regardless of its orientation. The four vertices of this rectangle are computed and rounded to integer pixel coordinates and are visualized as the green quadrilateral in the figure. The robot position used by the control policy is defined as the centroid of this quadrilateral, obtained by averaging the coordinates of its four vertices in image space, and returned as a two‑dimensional pixel vector, while a separate binary mask of all obstacle regions is generated by filling the contours associated with obstacles to provide a per‑pixel occupancy map aligned with the original camera frame.

\clearpage
\section{Height Regulation via PD Control}
To achieve 3D navigation, the control architecture assigns planar guidance and vertical regulation to separate control modules. The trained navigation policy dictates the yaw angle $\gamma$ to guide motion in the $x$-$y$ plane, while a Proportional-Derivative (PD) controller modulates the pitch angle $\beta$ to maintain the target altitude.

However, as indicated in Equation~(\ref{eq:3d_dynamics}), the planar and vertical dynamics are coupled through $\beta$. Specifically, the magnitude and direction of the planar velocity depend on $\sin \beta$. If $\sin \beta < 0$, the velocity vector in the $x$-$y$ plane opposes the direction defined by $\gamma$, resulting in reversed motion. To ensure the robot consistently advances along the heading generated by the navigation policy, we constrain $\beta$ to the interval $[0, \pi]$. This constraint guarantees that $\sin \beta \geq 0$, preserving the intended planar directionality, while allowing $\cos \beta$ to span $[-1, 1]$, thereby providing full authority over vertical actuation.

The vertical control loop operates by first minimizing the altitude error, $e_{z_t}$, defined as the deviation of the current robot height $z_{\text{robot}, t}$ from the target $z_\text{target}$:
\begin{equation}
e_{z_t} = z_\text{target} - z_{\text{robot}, t}
\end{equation}
Subsequently, the PD controller computes the required vertical velocity, $\hat{v}_z$:
\begin{equation}
\hat{v}_z = K_p e_{z_t} + K_d(e_{z_t} - e_{z_{t-1}})
\end{equation}
where the sampling interval is absorbed into the discrete derivative gain $K_d$.
To ensure physical feasibility, $\hat{v}_z$ is saturated within the dynamic limits derived from Equation~(\ref{eq:3d_dynamics}), specifically $[-\mu f - v_g, \mu f - v_g]$. Finally, the control command for the pitch angle is obtained by inverting the vertical dynamics:
\begin{equation}
\beta = \cos^{-1} \left( \frac{\hat{v}_z + v_g}{\mu f } \right)
\end{equation}

\clearpage
\section{Training Dataset}
We utilize a large-scale dataset for training. As illustrated in Fig.~\ref{fig:sup_dataset}, the original dataset is initially generated for the development of microswarm navigation strategies and comprises 43,133 items. Each item consists of an artificial vascular channel and a simulated microswarm. These artificial vascular channels are generated based on key characteristics of real vascular environments. Consequently, the dataset includes channel environments with varying diameters, branched channels, and open-sided channels with curved boundaries.

We further process the original dataset as follows. Specifically, we remove the simulated microswarms to obtain a set of isolated vessel environments. As the processed dataset contains duplicate entries, we eliminate these duplicates, resulting in a refined dataset of 14,753 unique instances. We then randomly partition this dataset into a training set (13,278 environments, 90 percent) and a test set (1,475 environments, 10 percent). During training (see Fig.~\ref{fig:sup_dataset}(C)), the microrobot's initial positions and targets are randomly assigned throughout the entire vascular space for each environment.

\clearpage
\section{Extended Discussion on Statistical Analysis}

Reinforcement learning results are known to exhibit substantial variability across independent runs, especially in the few-seed regime that is common in practice. Following prior methodological recommendations for deep RL evaluation~\cite{drl_statistical_2021nips}, we treat performance estimates obtained from a finite number of independent runs as random quantities and therefore report interval estimates, rather than relying only on point estimates or mean $\pm$ standard deviation. In this context, 95\% confidence intervals are used to communicate the uncertainty of the estimated mean across independent seeds, which is more directly relevant for assessing the stability and reliability of the reported performance differences. Specifically, the 95\% CI is computed as
\begin{equation}
    95\% \text{ CI} \approx \mu \pm 1.96 \frac{\sigma}{\sqrt{n}}
    \label{eq:95CI}
\end{equation}
where $\mu$ is the sample mean, $\sigma$ is the sample standard deviation across $n$ independent runs, and 1.96 is the critical value for a two-sided 95\% CI under a normal approximation. The SD quantifies variability across independent runs, whereas the 95\% CI quantifies uncertainty in the estimated mean. Because these quantities answer different questions, both may be reported when appropriate.

\subsection{The Unit of Statistical Analysis}

Here we provide a summary of the statistical analysis.
\begin{itemize}
    \item For time-related comparisons, such as those shown in Fig.~3(A)-(C), we compare hyperparameter settings at the run level rather than treating per-iteration timings as independent observations. Specifically, run-level comparison means that, for each independent run, we first average the performance over training iterations to obtain a run-level performance measure, and then compare these run-level averages across hyperparameter settings. This choice is important because the data collection time of individual training iterations can vary substantially across runs. A key reason is that the reset frequency changes over the course of training. In the early stages, resets occur more frequently because the randomly initialized policy fails more often. As training progresses and the policy converges, the success rate improves, resulting in fewer resets. Consequently, variation in reset frequency introduces noticeable fluctuations in data collection time across training iterations. We therefore conduct comparisons at the run level, where the results are relatively stable and more faithfully reflect performance differences under different hyperparameter settings.

    \item For reward-related statistics, we also compare hyperparameter settings at the run level. To compare final returns, we first average the returns over the last 50 iterations within each independent run and then compare these run-level averages across hyperparameter settings. This procedure provides a more reliable estimate of performance for each run. As shown in Fig.~3(E), the training curves have largely converged by 1000 training iterations; averaging over the last 50 iterations therefore reduces the influence of iteration-level random fluctuations on the overall statistical results. For training-curve comparisons, we perform statistical analysis at each training iteration across independent runs. Specifically, for each hyperparameter setting, we compute the mean performance and the 95\% confidence interval across runs at every iteration, thereby obtaining an average training curve with 95\% confidence intervals.
\end{itemize}

\subsection{The Choice of the Number of Independent Runs}

To make our statistical choices explicit, we conduct an additional sensitivity analysis for representative time-related and reward-related results shown in Figs.~3 and~4. First, we compute parametric 95\% CIs using Equation~\ref{eq:95CI} based on either 5 or 10 independent runs, in order to assess the sensitivity of the summary to the number of runs included. Second, using 10 independent runs, we compare the parametric 95\% CI with a percentile bootstrap 95\% CI to assess sensitivity to the interval-estimation method itself. Specifically, we resample the run-level means with replacement 1,000 times and recompute the mean for each bootstrap sample; the 2.5th and 97.5th percentiles of the resulting bootstrap distribution are taken as the lower and upper bounds of the 95\% CI, respectively. As shown in Fig.~\ref{fig:statistical_analysis}, the close agreement among these alternatives suggests that the main results are robust to both the number of runs included and the choice of interval-estimation procedure.

\clearpage
\section{Statistical Analysis of the Real-world Trajectories Under Long-horizon Navigation}

To further assess sim-to-real performance in real-world settings, we perform long-horizon navigation experiments in two representative environments. In a circular channel, four target points are defined, and the microrobot repeatedly navigates around the loop by sequentially reaching these points. In a branched channel, two target points are defined, and the microrobot repeatedly travels back and forth between them. Each trajectory group is repeated 15 times, resulting in six sets of repeated trials. The long-horizon experiments last approximately 37 min in the circular channel and 31 min in the branched channel.

Extended Data Fig. 1(C)(D) shows that the microrobot remained within the safe region and preserved a consistent clearance from obstacles across repeated trials. The small SDs across repeated trials indicate low trial-to-trial variability. We quantify performance using success rate, completion time, and path efficiency. The microrobot achieves a 100\% success rate in all six test cases, and the completion time is stable across repeated trials. Path efficiency is defined as the ratio between the straight-line distance and the actual traveled distance, with values closer to 1 indicating more efficient navigation. Across all test cases, the policy generates efficient trajectories, with slightly higher efficiency in scenario 1, likely owing to reduced interference from obstacles.

Notably, in S1-3, the microrobot occasionally performs a full in-place rotation before continuing its motion, while remaining at nearly the same position. This behaviour is likely the main reason for this group's relatively lower path efficiency, as the extra rotational adjustment increases the actual travel cost without advancing the robot toward the target.  This behaviour is likely caused by the angular output constraint $[-\pi, \pi)$, which introduces a discontinuity when switching between $-\pi$ and $\pi$. Future work will address this limitation to achieve smoother orientation transitions.

\clearpage

\section{Navigation in Complex Tasks}

For complex, large-scale tasks, a purely local reactive policy may be inadequate. Our framework can be readily extended to a hierarchical architecture, in which a high-level planner first computes a kinematic path from scene-level information and subsequently decomposes it into a sequence of local subgoals within the current field of view. The policy developed in this work then acts as a low-level controller, performing real-time navigation from local observations. Notably, the global path serves as guidance rather than a trajectory to be followed exactly, allowing the low-level controller to retain the key properties of the proposed framework, including local obstacle avoidance, robustness to external disturbances, and adaptation to dynamic obstacles. The learned policy should therefore be regarded as a local navigation module that can be coupled with a global planner for long-horizon tasks in larger and more complex environments.

As shown in Supplementary Fig.~\ref{fig:hierarchical_nav}, we evaluate this hierarchical strategy in vascular channels extracted from a publicly available retinal fundus image database~\cite{dataset_eye_2017}. In contrast to Fig. 2(C)(D), where navigation is restricted to a selected region of interest, here we consider the entire vascular network. As a proof of concept, we use A* as the high-level planner to generate a global kinematic path. This path is then used to determine receptive-field transitions and to define a local subgoal within each receptive field. The robot subsequently executes local navigation with the trained policy, progressively advancing towards the global target.

Future work will investigate reliable global map construction from real clinical measurements in conjunction with high-level path planning. Local-global matching and automated strategies for receptive-field transition also merit further study.

\clearpage
\section*{Supplementary Figures}

\begin{figure}[htbp]
    \centering
    \includegraphics[width=\linewidth, page=1, trim = 0 0 0 0, clip]{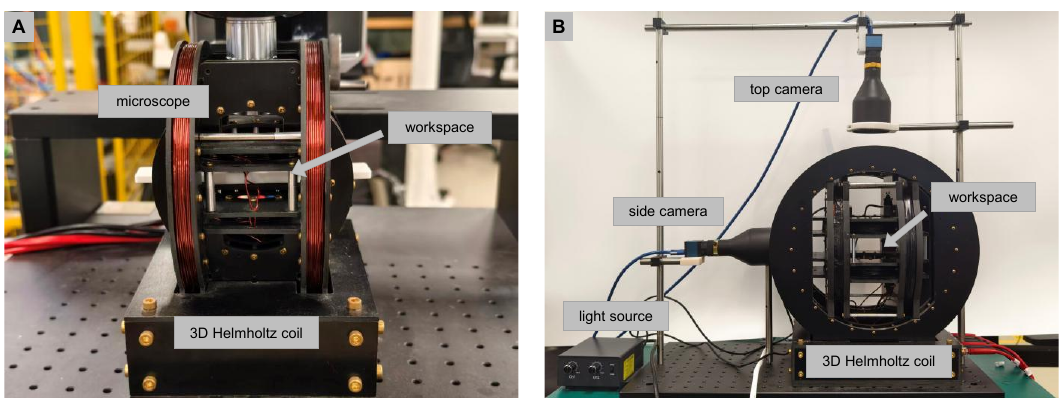}
    \caption{\textbf{Illustration of the hardware setup when deploying the trained policy in real-world scenarios.} (A) The 3D Helmholtz coil system for actuating the pollen-based microparticle. (B) The 3D Helmholtz coil system for actuating the helical robot.}
    \label{fig:hardware}
\end{figure}

\clearpage

\begin{figure}[htbp]
    \centering
    \includegraphics[width=\linewidth, page=1, trim = 0 0 0 0, clip]{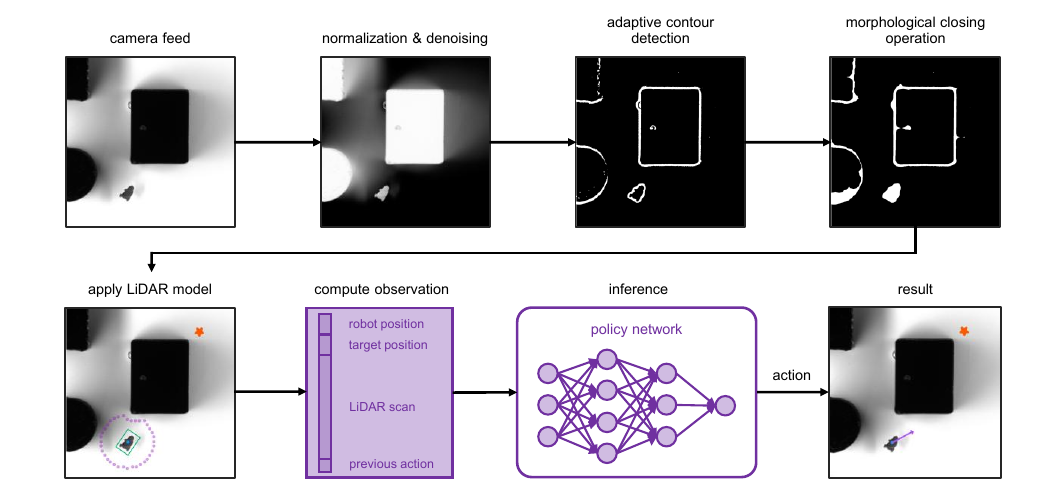}
    \caption{Overview of the workflow for deploying the trained policy in environments with real obstacles.}
    \label{fig:img_processing}
\end{figure}

\clearpage

\begin{figure}[htbp]
    \centering
    \includegraphics[width=\linewidth, page=1, trim = 0 0 0 0, clip]{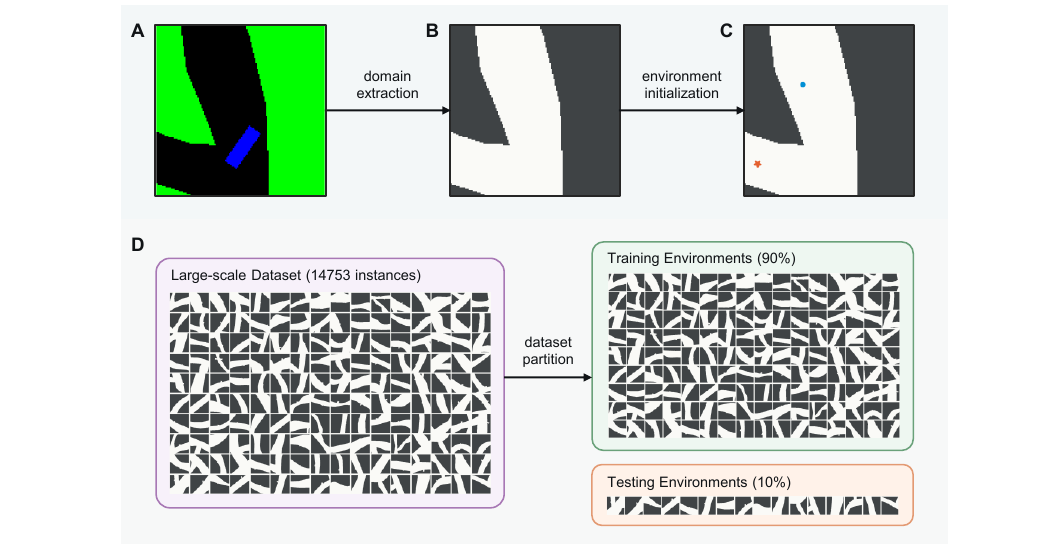}
    \caption{\textbf{Illustration of the large-scale dataset in the proposed parallel training framework.} (A) A sample from the original dataset, consisting of an artificial vascular channel and a simulated microswarm. The vascular space and obstacle space are indicated in black and green, respectively, while the blue square denotes the microswarm. (B) The extracted vascular channel obtained via image processing, where white and gray represent the vascular and obstacle spaces, respectively. (C) The corresponding training environment in simulation, with the blue circle and red star indicating the microrobot and the target position, respectively. (D) An overview of the constructed large-scale dataset, comprising 14,753 instances with diverse fundamental features that reflect real vascular environments. The dataset is partitioned into a training set (13,278 instances, 90\%) and a testing set (1,475 instances, 10\%).}
    \label{fig:sup_dataset}
\end{figure}

\clearpage

\begin{figure}[htbp]
    \centering
    \includegraphics[width=\linewidth, page=1, trim = 0 0 0 0, clip]{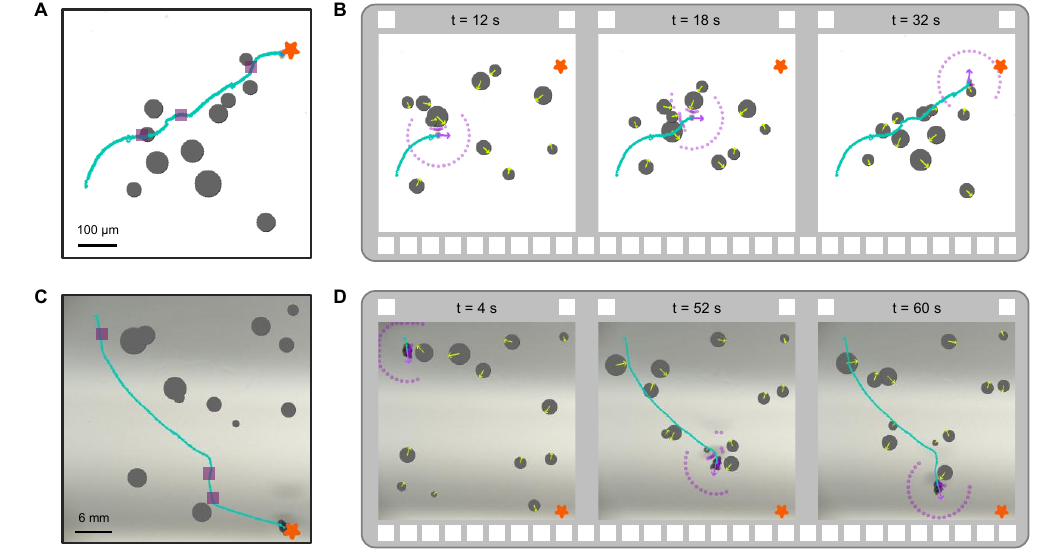}
    \caption{\textbf{Additional results for navigation across dynamic obstacles.} (A) Navigation of the pollen-based microparticle through dynamic obstacles; purple squares denote key frames along the trajectory. (B) Key frames for collision avoidance; yellow arrows indicate obstacle motion, and the purple arrow indicates the action generated by the policy network. (C) Navigation of the helical microrobot through dynamic obstacles. (D) Key frames for collision avoidance.}
    \label{fig:2d_obstacle_additional}
\end{figure}

\clearpage

\begin{figure}[htbp]
    \centering
    \includegraphics[width=\linewidth, page=1, trim = 0 0 0 0, clip]{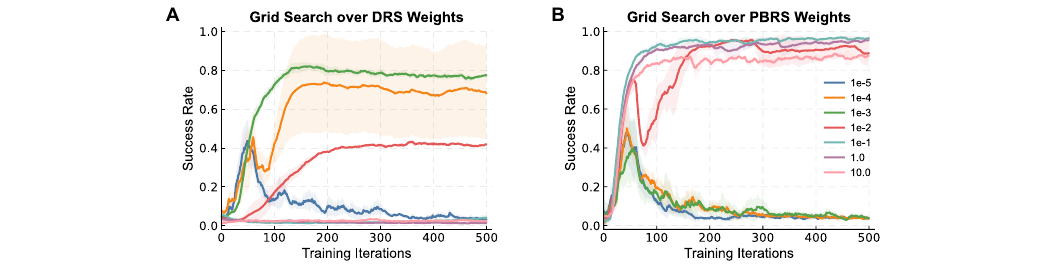}
    \caption{\textbf{Grid search results for determining the optimal weights of DRS and PBRS.} For each parameter setting, ten independent runs are conducted; solid lines indicate the mean performance and shaded areas denote the 95\% confidence interval across runs. To enhance visual clarity, the curves are smoothed from the raw experimental trajectories. (A) Results for DRS, where the optimal weight lies in the range between $10^{-4}$ and $10^{-3}$. (B) Results for PBRS, where the optimal weight lies in the range between $0.1$ and $1$. PBRS appears less sensitive to the choice of weight than DRS.}
    \label{fig:sup_rew_grid}
\end{figure}

\clearpage

\begin{figure}[htbp]
    \centering
    \includegraphics[width=\linewidth, page=1, trim = 0 0 0 0, clip]{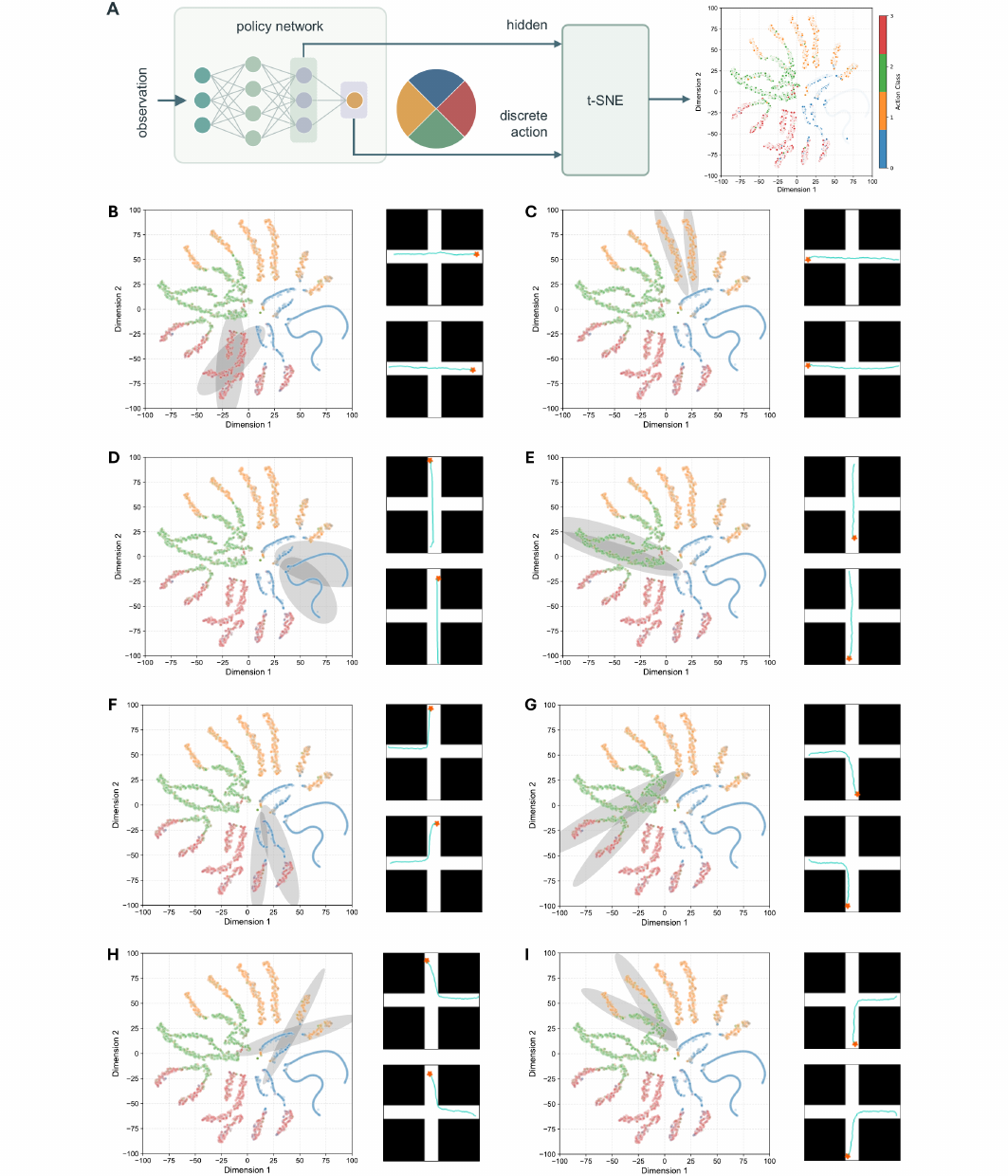}
    \caption{\textbf{More details of the t-SNE visualization.} (Additional details for the t-SNE visualization in Figure~2(C).) (A) Overview of the overall procedure. (B)--(I) Illustrations of the sampled trajectories and their corresponding data points in the visualization (highlighted by shaded regions). For clarity, a cross-shaped channel is used to generate trajectories. Two trajectories are sampled for each scenario.}
    \label{fig:sup_tsne_details}
\end{figure}

\clearpage

\begin{figure}[htbp]
    \centering
    \includegraphics[width=\linewidth, page=1, trim = 0 0 0 0, clip]{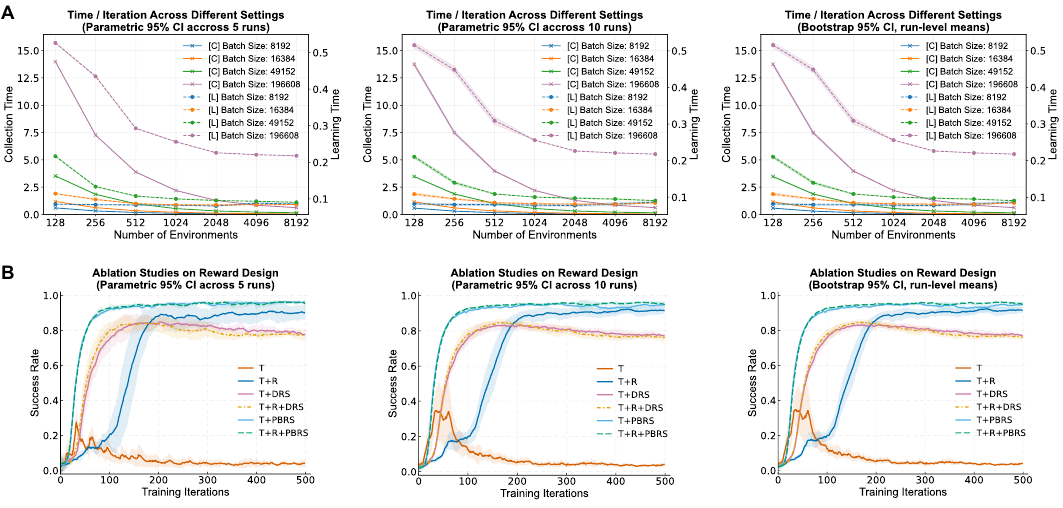}
    \caption{
    \textbf{Sensitivity analysis of statistical summaries across different settings.} We present two representative results: one based on a time-related statistic and the other on a reward-related statistic. For each case, we compare parametric
    95\% confidence intervals (CIs) computed from run-level means over 5 and 10 independent runs, together with bootstrap
    95\% CIs. The main conclusions remain consistent across these comparisons for both results. (A) Sensitivity analysis of the training time per iteration under different hyperparameter settings (corresponding to Fig.~3(B)). (B) Sensitivity analysis of the ablation study of the TSR framework (corresponding to Fig.~4(A)).
    }
    \label{fig:statistical_analysis}
\end{figure}

\clearpage

\begin{figure}[htbp]
    \centering
    \includegraphics[width=0.75\linewidth, trim = 0 0 0 0, clip]{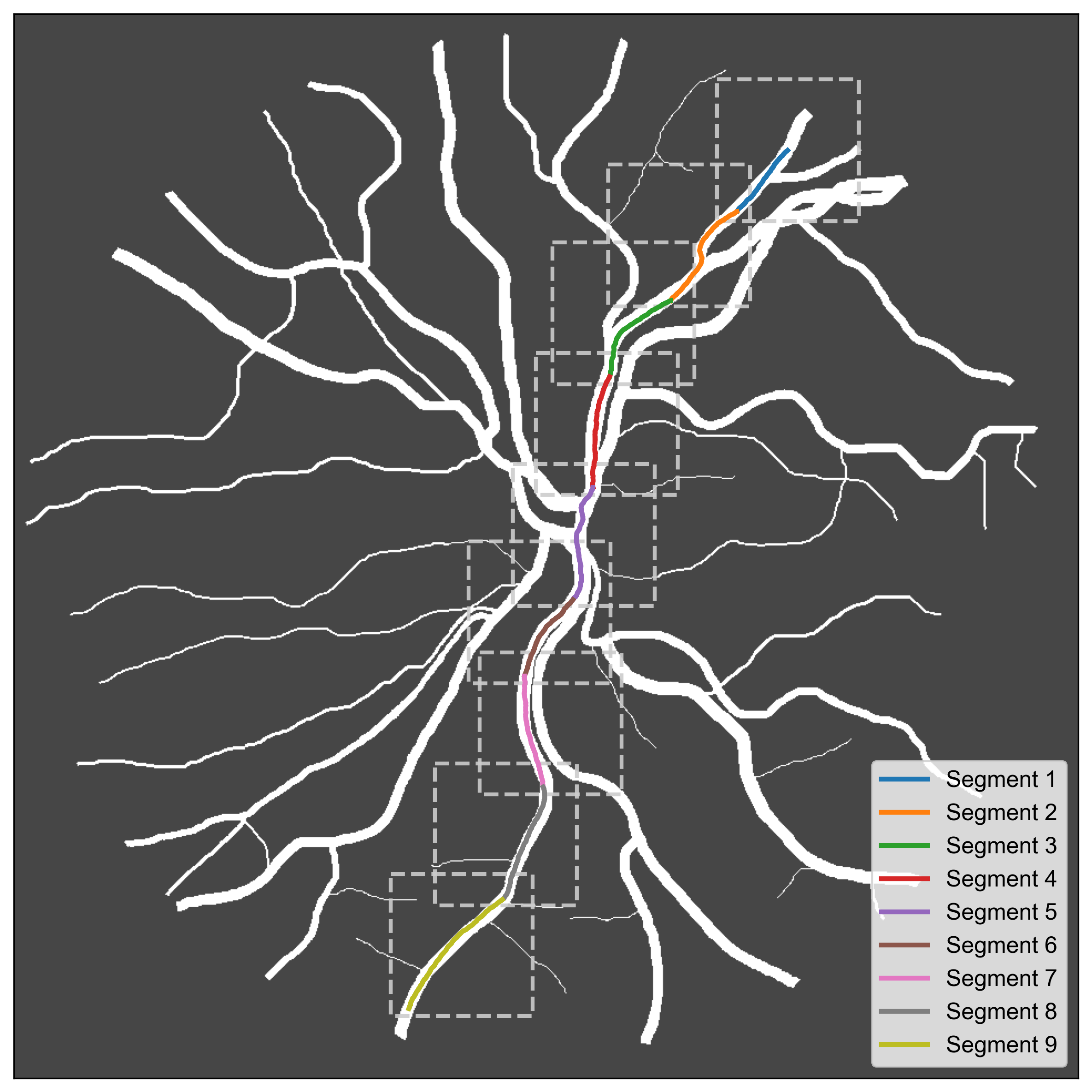}
    \caption{
    \textbf{Illustration of the hierarchical navigation in complex tasks.} We utilize a hierarchical navigation framework to guide the robot from the global start to the global target through a sequence of receptive fields. Specifically, a high-level global path planner first generates a coarse reference trajectory from a kinematic perspective. Based on this trajectory, we then determine how the receptive field moves and assign a subgoal to each receptive field. Finally, within each receptive field, the robot performs local navigation using our trained policy, thereby progressively reaching the global goal.
    }
    \label{fig:hierarchical_nav}
\end{figure}

\clearpage

\begin{figure}[htbp]
    \centering
    \includegraphics[width=\textwidth, trim = 0 0 0 0, clip]{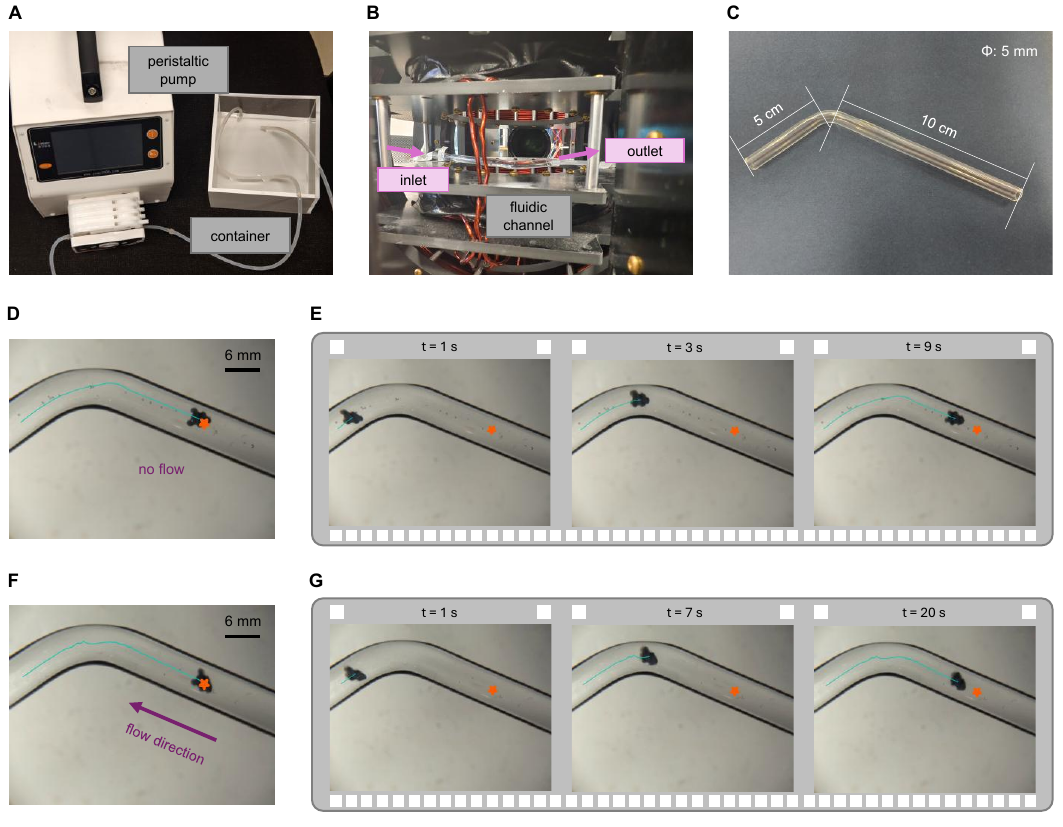}
    \caption{
    \textbf{Sim-to-real validation in a fluid-flow environment.} (A) Additional hardware setup for experiments under flow conditions; a peristaltic pump is used to generate the flow. (B) Schematic illustration of the fluidic channel in the workspace. (C) Dimensions of the fluidic channel. (D) Navigation of the helical microrobot in the absence of flow, serving as the control condition. (E) Representative key frames of navigation without flow. (F) Navigation of the helical microrobot under a flow speed of 2~mm/s. (G) Representative key frames of navigation in the fluid-flow environment.
    }
    \label{fig:flow_env}
\end{figure}

\clearpage
\section*{Supplementary Tables}
\begin{table}[htbp]
    \centering
    \renewcommand{\arraystretch}{1.2}
    \caption{\textbf{Hyperparameters of the PPO Algorithm}}
    \label{tbl:ppo_hyperparameters}
    \vspace{4mm}
    \begin{tabular}{l|l}
        \hline
        Hyper-parameter & Value \\
        \hline
        Batch size & 196,608 \\
        Mini-batch size & 49,152 \\
        Number of epochs & 5 \\
        clip range & 0.2 \\
        Entropy coefficient & 0.01 \\
        Discount factor ($\gamma$) & 0.99 \\
        GAE-$\lambda$ & 0.95 \\
        Learning rate & adaptive \\
        Desired KL-divergence & 0.01\\
        \hline
    \end{tabular}
\end{table}

\clearpage
\begin{table}[htbp]
    \centering
    \renewcommand{\arraystretch}{1.2}
    \caption{\textbf{The network structure.}}
    \label{tbl:nn_param}
    \vspace{4mm}
    \begin{tabular}{l|l}
        \hline
        Network & Structure \\
        \hline
        Actor & (1) Linear (In = 77, Out = 256, Bias = True) \\
              & (2) ELU (alpha = 1.0) \\
              & (3) Linear (In = 256, Out = 128, Bias = True) \\
              & (4) ELU (alpha = 1.0) \\
              & (5) Linear (In = 128, Out = 16, Bias = True) \\
              & (6) ELU (alpha = 1.0) \\
              & (7) Linear (In = 16, Out = 1, Bias = True) \\
        \hline
        Critic& (1) Linear (In = 77, Out = 256, Bias = True) \\
              & (2) ELU (alpha = 1.0) \\
              & (3) Linear (In = 256, Out = 128, Bias = True) \\
              & (4) ELU (alpha = 1.0) \\
              & (5) Linear (In = 128, Out = 16, Bias = True) \\
              & (6) ELU (alpha = 1.0) \\
              & (7) Linear (In = 16, Out = 1, Bias = True) \\
        \hline
    \end{tabular}
\end{table}

\clearpage
\begin{table}[htbp]
    \centering
    \renewcommand{\arraystretch}{1.2}
    \caption{\textbf{The reward coefficients.}}
    \label{tbl:rew_coef}
    \vspace{4mm}
      \begin{adjustbox}{max width=\textwidth}
  \begin{threeparttable}
    \begin{tabular}{l|l|l|l}
        \hline
        Classification & Term & Equation & Coefficient \\
        \hline
        Task reward & destination & $ \mathbb{I}\{ \|\bm{p}_{\text{robot}, t} - \bm{p}_\text{target} \|_2 \leq \epsilon\}$ & 10 \\
                    & feasibility & $\mathbb{I}\{\bm{p}_{\text{robot}, t} \notin \mathcal{V} \}$ & -10 \\
        \hline
        Shaping reward & potential-based shaping & $d_{t-1} - d_{t}$ & 0.5 \\
        \hline
        Regularization reward & action smoothness & $\|\bm{a}_t - \bm{a}_{t-1} \|_2$ & -0.01 \\
                              & obstacle safety & $\frac{1}{N} \sum_{i=1}^N \log \| \bm{m}_i - \bm{p}_{\text{robot}, t} \|_2$ & 0.001\\
        \hline
    \end{tabular}
    \begin{tablenotes}
\small
\vspace{1mm}
\item Notes: $d_t = \| \bm{p}_{\text{robot}, t} - \bm{p}_\text{target} \|_2$
\end{tablenotes}

\end{threeparttable}

  \end{adjustbox}
\end{table}

\clearpage
\begin{table}[htbp]
  \centering
    \renewcommand{\arraystretch}{1.2}
    \caption{\textbf{Ablation study results for regularization rewards.}}
    \label{tbl:sup_data_abl_reg_rew}
    \vspace{4mm}

  \begin{adjustbox}{max width=\textwidth}
  \begin{threeparttable}
  \begin{tabular}{c|cc|cc|cc|cc}
    \hline
    & \multicolumn{2}{c|}{Scene 1}
    & \multicolumn{2}{c|}{Scene 2}
    & \multicolumn{2}{c|}{Scene 3}
    & \multicolumn{2}{c}{Scene 4} \\
    \cline{2-9}
    & Safety ($\uparrow$) & Smoothness ($\downarrow$) 
    & Safety ($\uparrow$) & Smoothness ($\downarrow$)
    & Safety ($\uparrow$) & Smoothness ($\downarrow$)
    & Safety ($\uparrow$) & Smoothness ($\downarrow$) \\
    \hline
    with reg. & 5.915 & 0.288 & 5.743 & 0.281 & 6.266 & 0.299 & 5.936 & 0.273 \\
    Trial 1 & $\pm$ 0.195 & $\pm$ 0.061 & $\pm$ 0.236 & $\pm$ 0.052 & $\pm$ 0.204 & $\pm$ 0.070 & $\pm$ 0.106 & $\pm$ 0.049 \\
    \hline
    with reg. & 5.925 & 0.267 & 5.937 & 0.258 & 6.335 & 0.277 & 5.863 & 0.252 \\
    Trial 2 & $\pm$ 0.189 & $\pm$ 0.054 & $\pm$ 0.145 & $\pm$ 0.044 & $\pm$ 0.169 & $\pm$ 0.060 & $\pm$ 0.094 & $\pm$ 0.040 \\
    \hline
    with reg. & 5.799 & 0.299 & 5.927 & 0.257 & 6.327 & 0.301 & 5.754 & 0.283 \\
    Trial 3 & $\pm$ 0.217 & $\pm$ 0.202 & $\pm$ 0.136 & $\pm$ 0.045 & $\pm$ 0.186 & $\pm$ 0.053 & $\pm$ 0.104 & $\pm$ 0.179 \\
    \hline
    with reg. & 5.935 & 0.294 & 5.899 & 0.294 & 6.340 & 0.299 & 5.980 & 0.282 \\
    Trial 4 & $\pm$ 0.183 & $\pm$ 0.048 & $\pm$ 0.191 & $\pm$ 0.219 & $\pm$ 0.163 & $\pm$ 0.053 & $\pm$ 0.070 & $\pm$ 0.033 \\
    \hline
    with reg. & 5.986 & 0.284 & 5.793 & 0.273 & 6.337 & 0.276 & 5.977 & 0.267 \\
    Trial 5 & $\pm$ 0.179 & $\pm$ 0.062 & $\pm$ 0.194 & $\pm$ 0.053 & $\pm$ 0.178 & $\pm$ 0.060 & $\pm$ 0.142 & $\pm$ 0.049 \\
    \hline
    no reg. & 5.807 & 0.444 & 5.692 & 0.423 & 6.173 & 0.460 & 5.709 & 0.431 \\
    Trial 1 & $\pm$ 0.180 & $\pm$ 0.071 & $\pm$ 0.188 & $\pm$ 0.055 & $\pm$ 0.207 & $\pm$ 0.080 & $\pm$ 0.157 & $\pm$ 0.051 \\
    \hline
    no reg. & 5.789 & 0.465 & 5.586 & 0.439 & 6.036 & 0.477 & 5.717 & 0.436 \\
    Trial 2 & $\pm$ 0.199 & $\pm$ 0.074 & $\pm$ 0.228 & $\pm$ 0.064 & $\pm$ 0.259 & $\pm$ 0.074 & $\pm$ 0.105 & $\pm$ 0.057 \\
    \hline
    no reg. & 5.706 & 0.414 & 5.755 & 0.387 & 6.271 & 0.430 & 5.805 & 0.381 \\
    Trial 3 & $\pm$ 0.169 & $\pm$ 0.059 & $\pm$ 0.191 & $\pm$ 0.047 & $\pm$ 0.195 & $\pm$ 0.064 & $\pm$ 0.132 & $\pm$ 0.045 \\
    \hline
    no reg. & 5.899 & 0.408 & 5.802 & 0.385 & 6.279 & 0.418 & 5.774 & 0.389 \\
    Trial 4 & $\pm$ 0.186 & $\pm$ 0.053 & $\pm$ 0.172 & $\pm$ 0.046 & $\pm$ 0.165 & $\pm$ 0.052 & $\pm$ 0.137 & $\pm$ 0.038 \\
    \hline
    no reg. & 5.749 & 0.440 & 5.755 & 0.421 & 6.119 & 0.445 & 5.813 & 0.411 \\
    Trial 5 & $\pm$ 0.232 & $\pm$ 0.096 & $\pm$ 0.242 & $\pm$ 0.092 & $\pm$ 0.176 & $\pm$ 0.096 & $\pm$ 0.102 & $\pm$ 0.075 \\
    \hline
  \end{tabular}
\begin{tablenotes}
\small
\vspace{1mm}
\item Notes: (1) \(\uparrow\) denotes that higher values are better, while \(\downarrow\) denotes that lower values are better. (2) For each entry in the table, the statistics are computed from 10,000 trajectories collected by executing the corresponding policy in the corresponding scenario, where the initial and goal states of each trajectory are randomly sampled. (3) The collision-avoidance metric is computed using only scan points within 10 units of the robot, emphasizing near-obstacle interactions and mitigating the undue influence of distant points.
\end{tablenotes}

\end{threeparttable}

  \end{adjustbox}
\end{table}

\clearpage

\section*{Supplementary Movies}

\paragraph{Movie S1.} {Illustration of the ultra-fast learning process.}

\paragraph{Movie S2.} {Zero-shot adaptation to unseen scenarios.}

\paragraph{Movie S3.} {Zero-shot sim-to-real adaptation for different kinds of microrobots in channel environments.}

\paragraph{Movie S4.} {Zero-shot sim-to-real adaptation for different kinds of microrobots in dynamic environments.}

\paragraph{Movie S5.} {Autonomous navigation in 3D environments with physical obstacles.}

\paragraph{Movie S6.} {3D Autonomous Navigation in a Human Brain Vasculature Model.}

\paragraph{Movie S7.} {Zero-shot sim-to-real adaptation in fluid flow environments.}

\paragraph{Movie S8.} {Long-horizon sim-to-real validation over multiple repeated trajectories.}

\hypersetup{pageanchor=true}

\end{document}